\documentclass[10pt,journal,compsoc]{IEEEtran}

\makeatletter
\def\endthebibliography{%
	\def\@noitemerr{\@latex@warning{Empty `thebibliography' environment}}%
	\endlist
}
\makeatother

\ifCLASSOPTIONcompsoc
  \usepackage[nocompress]{cite}
\else
  \usepackage{cite}
\fi

\usepackage{hyperref}       % hyperlinks
\hypersetup{colorlinks=true,urlcolor=black}
\usepackage{url}            % simple URL typesetting
\usepackage{booktabs}       % professional-quality tables
\usepackage{amsfonts}       % blackboard math symbols
\usepackage{nicefrac}       % compact symbols for 1/2, etc.
\usepackage{microtype}      % microtypography
\usepackage{algorithm}
\usepackage{algorithmic}
\usepackage{graphicx}
\usepackage{epsfig}
\usepackage{amsmath}
\usepackage{amssymb}
\usepackage{amsthm}
\usepackage{subfigure}
\usepackage{booktabs} % for professional tables
\usepackage{multirow}
\usepackage{etoolbox}
\usepackage{bbm}
\usepackage{sail}
\usepackage{color}
\usepackage{footnote}
\usepackage{arydshln}
\usepackage{enumitem}
\usepackage{times}
\usepackage{ragged2e}
\usepackage{xspace}
\usepackage{threeparttable}
\usepackage[export]{adjustbox}

\def\shortname{REO\xspace}
\def\name{Calibration-free Spatial Transformation\xspace}
\def\lowername{calibration-free spatial transformation\xspace}

\begin{document}
%
% TODOLIST: 
% 1. 根据审稿人意见，不要claim calibration-free，换成其他说法，比如 implicit spatial representation learning （ISRL）
% 2. 根据已有v2版本的验证实验，提升速度 20x
% 3. 分开3d texture cues 的head 以及实际任务的head，这样不会导致两个任务相互影响
% 4. 分开head之后，分割任务的可以加上更多层的conv或者mlp，提升任务性能
% 5. 提升cam-only 的性能，使其comparable
% 6. 增加 radar实验，后面就cam, cam+lidar, cam+radar, cam+lidar+radar这样去比较
% 7. 补充relatework里面lidar 的方法
% 8. 在semantickitti test 上提交结果

% 写作思路：compact spatial representation learning -> 降低标定信息的依赖，自动学习空间映射关系；->考虑速度方面，限制feature 大小-->为了进一步提升表征能力-->引入辅助任务2D任务；为了强化空间学习能力-->引入3D texture cues;

\title{Robust 3D Semantic Occupancy Prediction with Calibration-free Spatial Transformation}

\author{
	Zhuangwei Zhuang$^*$, Ziyin Wang$^*$, Sitao Chen$^*$, Lizhao Liu, Hui Luo,  Mingkui Tan$^\dagger$
	\IEEEcompsocitemizethanks{
        \IEEEcompsocthanksitem Zhuangwei Zhuang, Sitao Chen, Lizhao Liu and Mingkui Tan are with the School of Software Engineering, South China University of Technology. Zhuangwei Zhuang is also with Robosense Inc., ShenZhen, China. Mingkui Tan is also with the Pazhou Laboratory, Guangzhou, China. E-mail:\{z.zhuangwei, mechenst, selizhaoliu\}@mail.scut.edu.cn, mingkuitan@scut.edu.cn.
        \IEEEcompsocthanksitem Ziyin Wang is with Robosense Inc., ShenZhen, China. E-mail: wang2457@purdue.edu.
        \IEEEcompsocthanksitem Hui Luo is with the National Key Laboratory of Optical Field Manipulation Science and Technology, CAS, also with the Institute of Optics and Electronics, CAS, Chengdu, China. E-mail: luohui19@mails.ucas.ac.cn.
        
        % \IEEEcompsocthanksitem Yuanqing Li is with the Pazhou Laboratory, Guangzhou, China. E-mail: auyqli@scut.edu.cn.
		\IEEEcompsocthanksitem $^*$Authors contribute equally. $^\dagger$ Corresponding author.
	}
}

% in the abstract or keywords.
\IEEEtitleabstractindextext{%
\begin{abstract}
\justifying
3D semantic occupancy prediction, which seeks to provide accurate and comprehensive representations of environment scenes, is important to autonomous driving systems. For autonomous cars equipped with multi-camera and LiDAR, it is critical to aggregate multi-sensor information into a unified 3D space for accurate and robust predictions. Recent methods are mainly built on the 2D-to-3D transformation that relies on sensor calibration to project the 2D image information into the 3D space. These methods, however, suffer from two major limitations: First, they rely on accurate sensor calibration and are sensitive to the calibration noise, which limits their application in real complex environments. Second, the spatial transformation layers are computationally expensive and limit their running on an autonomous vehicle. In this work, we attempt to exploit a \textbf{R}obust and \textbf{E}fficient 3D semantic \textbf{O}ccupancy (\textbf{\shortname}) prediction scheme. To this end, we propose a calibration-free spatial transformation based on vanilla attention to implicitly model the spatial correspondence. In this way, we robustly project the 2D features to a predefined BEV plane without using sensor calibration as input. Then, we introduce 2D and 3D auxiliary training tasks to enhance the discrimination power of 2D backbones on spatial, semantic, and texture features. Last, we propose a query-based prediction scheme to efficiently generate large-scale fine-grained occupancy predictions. By fusing point clouds that provide complementary spatial information, our \shortname surpasses the existing methods by a large margin on three benchmarks, including OpenOccupancy, Occ3D-nuScenes, and SemanticKITTI Scene Completion. For instance, our \shortname achieves \textbf{19.8}$\times$ speedup compared to Co-Occ, with \textbf{1.1}\% improvements in geometry IoU on OpenOccupancy. Our code will be available at \url{https://github.com/ICEORY/REO}.
\end{abstract}

% Note that keywords are not normally used for peerreview papers.
\begin{IEEEkeywords}
Multi-Sensor Fusion, 3D Semantic Occupancy Prediction, Scene Understanding, Autonomous Driving.
\end{IEEEkeywords}}

% make the title area
\maketitle

% To allow for easy dual compilation without having to reenter the
% abstract/keywords data, the \IEEEtitleabstractindextext text will
% not be used in maketitle, but will appear (\ie, to be "transported")
% here as \IEEEdisplaynontitleabstractindextext when the compsoc
% or transmag modes are not selected <OR> if conference mode is selected
% - because all conference papers position the abstract like regular
% papers do.
\IEEEdisplaynontitleabstractindextext
% \IEEEdisplaynontitleabstractindextext has no effect when using
% compsoc or transmag under a non-conference mode.

% For peer review papers, you can put extra information on the cover
% page as needed:
% \ifCLASSOPTIONpeerreview
% \begin{center} \bfseries EDICS Category: 3-BBND \end{center}
% \fi
%
% For peerreview papers, this IEEEtran command inserts a page break and
% creates the second title. It will be ignored for other modes.
\IEEEpeerreviewmaketitle

\IEEEraisesectionheading{\section{Introduction}\label{sec:introduction}}
\IEEEPARstart{A}{ccurate} and comprehensive 3D perception of the surrounding environment is crucial for robotic systems such as autonomous drivings~\cite{liao2022kitti,sun2020scalability} and robotic navigation~\cite{chen2022learning,krantz2020beyond}. To improve the perception ability of autonomous cars, researchers have developed different perception tasks, including object detection~\cite{yang2018pixor,yin2021center}, semantic segmentation~\cite{zhuang2021perception,graham20183d,jaritz2020xmuda}, and panoptic segmentation~\cite{zhou2021panoptic,zhang2023lidar} in the 3D space. These tasks, however, fail to provide a comprehensive and coherent representation of the complex outdoor scenes~\cite{tian2023occ3d}. To address this issue, 3D semantic occupancy prediction, which provides the geometry and semantics of every voxel in the 3D space, has become a fundamental task for modern autonomous systems~\cite{hu2023planning,gu2023vip3d}.

\begin{figure}[t]
\centering
    \includegraphics[width=1.0\linewidth]{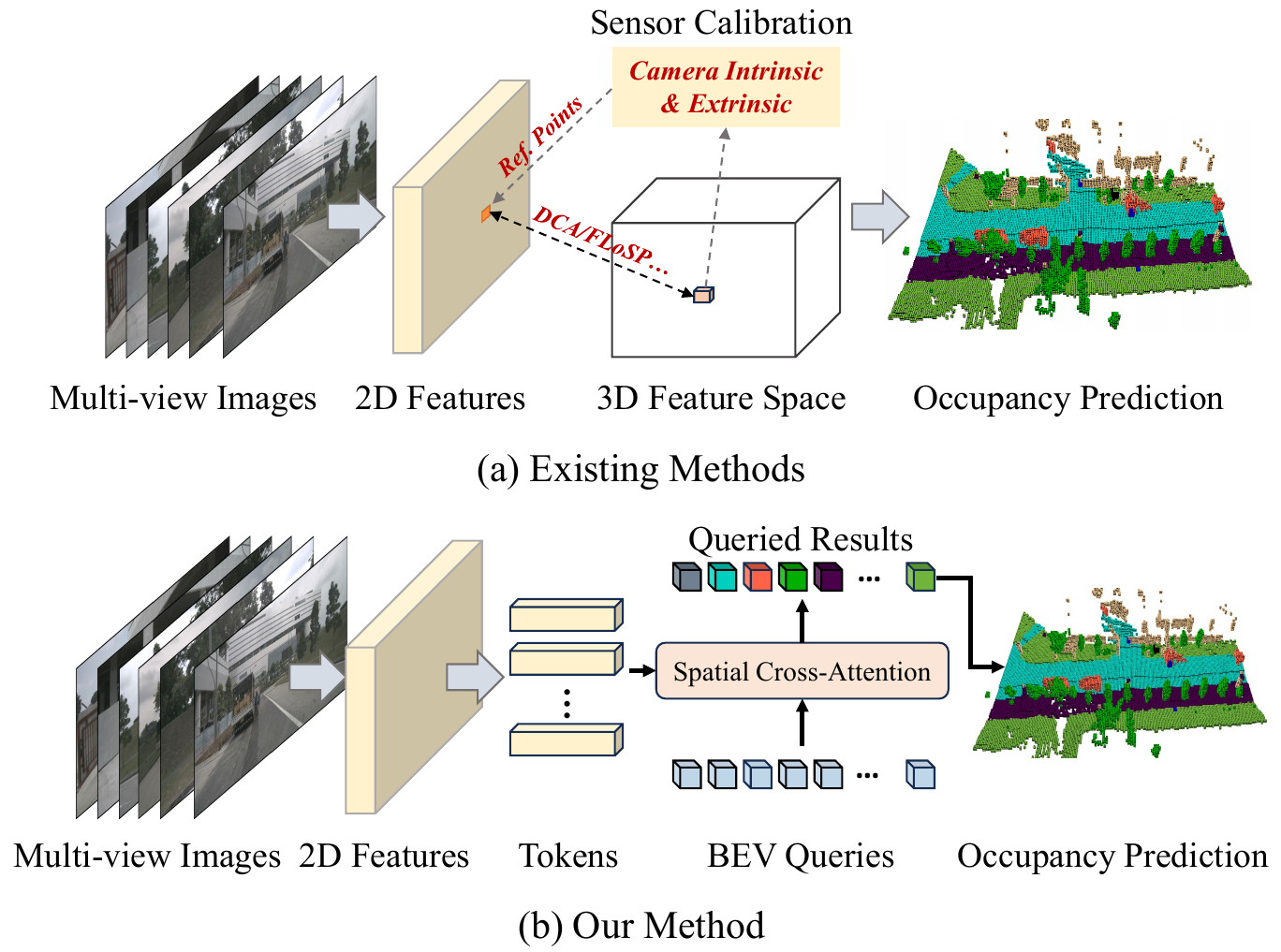}
    \caption{Comparisons of existing methods~\cite{li2022bevformer,li2023voxformer,huang2023tri,cao2022monoscene} and our \shortname. Unlike existing methods that rely on sensor calibration to compute the reference points of the 3D voxels, our \shortname directly models the 2D-to-3D spatial correspondence by attention scheme without using sensor calibration. We simplify the architecture diagram for better illustration.}
     \label{fig:cmp_bevformer}
\end{figure}

As autonomous cars are often equipped with multi-sensor (~\ie, multi-cameras and LiDARs), the core of 3D semantic occupancy prediction lies in how to effectively aggregate the features from 2D and 3D sensors in a unified 3D space to generate accurate 3D predictions. The past years have witnessed the great success of BEVFormer-like methods, which represent the multi-view camera features in a bird's-eye-view (BEV) plane using spatial transformation layers that consist of deformable cross-attention (DCA) and deformable self-attention (DSA) layers~\cite{li2022bevformer,tong2023scene,wei2023surroundocc}. Recent studies further improve the representation ability of deep models by aggregating the image features in a tri-perspective (TPV) plane~\cite{huang2023tri}. Despite their effectiveness, these methods, however, require calibration information (\ie, intrinsic or extrinsic of the sensors) to aggregate the features around the reference points from multi-view cameras(as shown in Figure~\ref{fig:cmp_bevformer}(a)). The coordinates of reference points may be inaccurate due to the calibration noise in real-life applications and may degrade the model performance of these methods~\cite{liu2022petr,jiang2022multi}\footnote{See discussions in Sec.~\ref{sec:discuss_calibration}.}. Moreover, limited by the 2D-to-3D projection scheme, these methods often obtain sparse features and require multi-resolution BEV/TPV queries to generate fine-grained occupancy predictions\footnote{See discussions in Sec.~\ref{sec:vis_bev_feat}.}. As a result, the computational costs of these methods are unaffordable for real-life applications~\cite{xu2025survey} (See Table~\ref{tab:dca_infer_speed}).

\begin{table}[t]
    \caption{Inference time of spatial transformation in existing methods. Both VoxFormer and TPVFormer use DSA+DCA to lift the image feature. MonoScene uses FLoSP to project the image features into 3D space. We report the results evaluated on SemanticKitti Scene Completion.}
    \centering
    \begin{tabular}{c|cc|c}
    \hline
    Methods  & Spatial Transform. & Total & Percentage (\%)\\
    \hline
    VoxFormer~\cite{li2023voxformer} & 72.75ms & 80.45ms & 90.42\% \\
    TPVFormer~\cite{huang2023tri} & 118.99ms & 172.96ms & 68.80\% \\
    MonoScene~\cite{cao2022monoscene} & 78.81ms & 145.26ms & 54.25\% \\
    \hline
    % \shortname (Ours) & 24.62ms & 44.26ms & 55.63\% \\
    \shortname (Ours) & 5.02ms & 44.26ms & 11.34\% \\
    \hline
    \end{tabular}
    
    \label{tab:dca_infer_speed}
\end{table}

In this work, we argue that the dependency on sensor calibration information can be eliminated, motivated by the fact that \textit{Humans can learn to reconstruct the physical world in the brain even without knowing the exact positions of eyes (a.k.a. sensor extrinsic parameters).} Moreover, inspired by~\cite{zhou2022cross}, which implicitly learns a mapping from multi-cameras into a 2D map-view representation using a cross-view attention mechanism, we further explore a \name paradigm that implicitly learns the 2D-to-3D spatial representation with vanilla cross-attention scheme.  Specifically, as shown in Figure~\ref{fig:cmp_bevformer}(b), given sampled voxels in the 3D space, we obtain the occupancy predictions directly from multi-view images via cross-attention modules. As the cross-attention scheme considers the full correlation between 2D features and 3D space, it is insensitive to the sensor calibration noise. Though the idea is straightforward, it suffers from two limitations. \textbf{First}, the computational costs and memory requirements are unaffordable due to the high-resolution nature of multi-view images and a vast number of queried voxels ($\sim 10^5$ on Occ3D-nuScenes). \textbf{Second}, it is difficult to dig out both the spatial correspondence and discriminative features from multi-view images without spatial prior from sensor calibration information~\cite{chen2022efficient}.

To address these issues, we exploit a Robust and Efficient Occupancy (\shortname) prediction scheme with calibration-free spatial transformation based on vanilla cross-attention. Specifically, we first extract the image features by a pre-trained 2D encoder and propose an image feature aggregation module to aggregate the features from different cameras. 
Then, we propose calibration-free spatial-transformation modules based on cross-attention to transform the image features into a compact BEV plane. Third, we lift the BEV features to voxel space and introduce a query-based prediction scheme to generate both geometry and semantic predictions efficiently. As it is difficult to learn both 2D-to-3D projection and semantic representations without spatial prior, we introduce auxiliary 2D and 3D training tasks to enhance the discrimination power of the image backbone on spatial, textural, and semantic features. The main contributions of this work are summarized as follows.

\begin{itemize}
    \item We find that the sensor calibration for multi-sensor fusion can be discarded during inference by introducing calibration-free spatial transformation (CST) modules based on the vanilla attention scheme. With the proposed CST, our \shortname is robust in calibration noise of sensor-calibration information (\ie, camera intrinsic and extrinsic parameters).  
    \item We introduce both 2D and 3D auxiliary training tasks, including depth regression, semantic segmentation, and texture reconstruction, to enhance the discrimination power of the image backbone and effectively improve the performance of the occupancy model.
    \item We propose a query-based prediction scheme to efficiently predict the geometry and semantic occupancy results of sampled queried voxels in 3D space. Moreover, compared to the state-of-the-art fusion-based method (\ie, Co-Occ), our \shortname achieves \textbf{19.8}$\times$ acceleration with \textbf{1.1}\% improvements in geometry IoU on OpenOccupancy.
    % \item We propose \shortname to effectively fuse the multi-sensor information and represent the 3D scene in the implicit feature space with 2D-3D reconstruction guidance.
    % \item The proposed \shortname is built upon the cross-attention modules and query-based mechanism, which provide a global perception of multi-sensor features and flexible extension for downstream tasks.
    % \item Our query-based token compression module and coarse-to-fine query scheme can effectively reduce the computational complexity of our method and make it available for real-life application. 
\end{itemize}
\begin{figure}[t]
\centering
    \includegraphics[width=\linewidth]{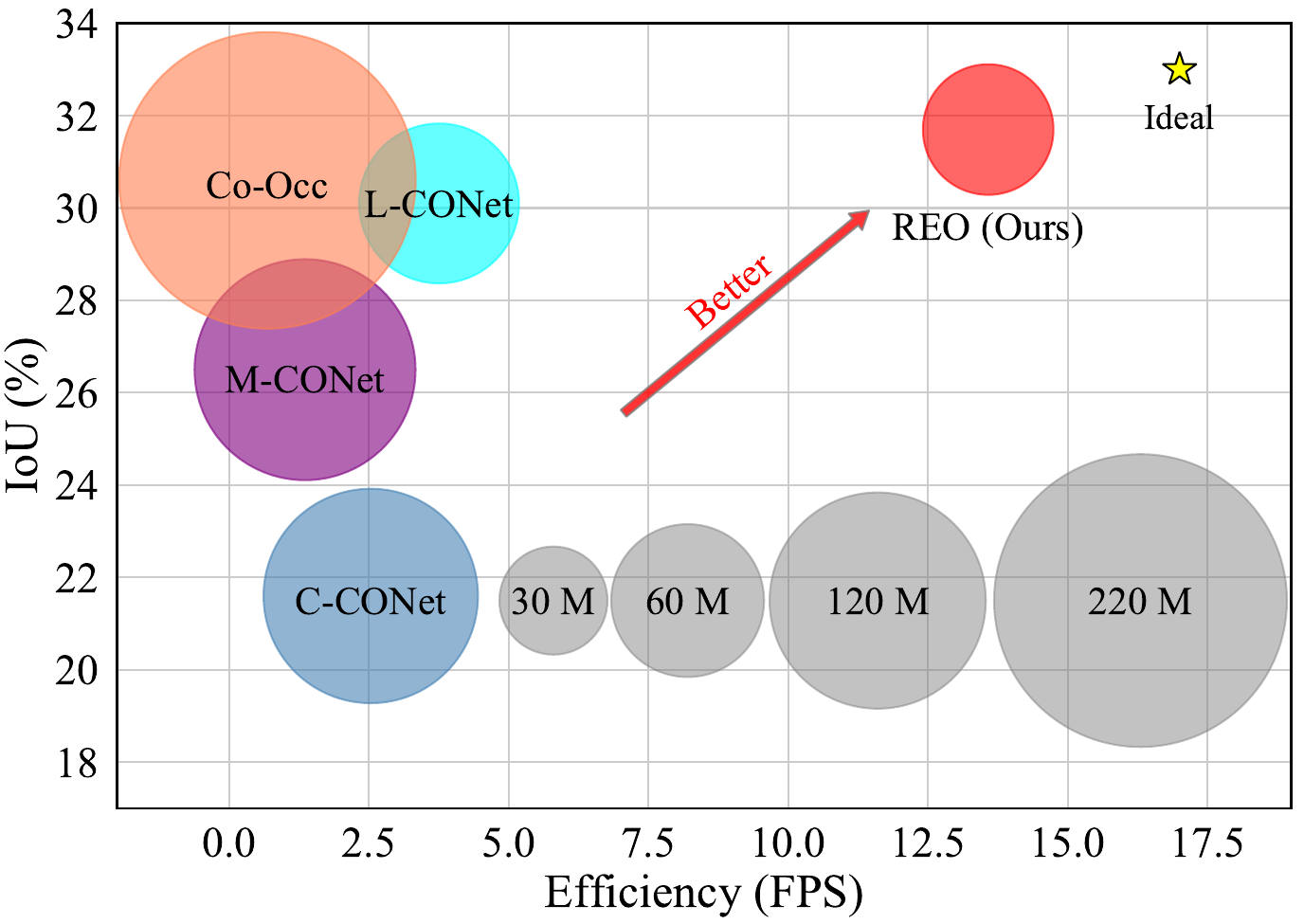}
    \caption{Comparisons of model efficiency and performance of different methods on OpenOccupancy.}
    \label{fig:cmp_efficient}
\end{figure}

\section{Related Work}
\noindent\textbf{Multi-sensor fusion based perception}
% \subsection{Multi-sensor Fusion}
As cameras and LiDAR can provide complementary information, many works~\cite{el2019rgb, meyer2019sensor, vora2020pointpainting, zhuang2021perception} aim to efficiently fuse the multi-sensor features for accurate and robust 3D perception. RGBAL~\cite{el2019rgb} converts RGB images to a polar-grid mapping representation and designs early and mid-level fusion architectures. EPMF~\cite{zhuang2021perception,tan2024epmf} projects the point clouds to the camera coordinate system using perspective projection and proposes a collaborative fusion scheme with perception-aware loss. 2DPASS~\cite{yan20222dpass} designs a multi-scale fusion-to-single knowledge distillation strategy to distill multi-modal knowledge to single point cloud modality. M-CONet~\cite{wang2023openoccupancy} constructs a LiDAR branch and a camera branch for point clouds and images, respectively, and projects queries to sample from 2D image features and 3D voxel features and fuses them through an adaptive fusion module. Recent works(\eg, DeepFusion~\cite{li2022deepfusion}, FUTR3D~\cite{chen2023futr3d}, UniSeg~\cite{liu2023uniseg}) attempt to mitigate the problem of feature misalignment by a learnable sensor fusion. Different from the existing fusion-based method, our \shortname is camera-centric and only uses a small number of voxelized point clouds, reducing the dependency on expensive high-resolution LiDAR. Besides, the multi-modality features are implicitly aligned via cross-attention modules without leveraging spatial prior from calibration information. 
% However, the above methods require strict alignment of cross-modal features. Although existing work (\eg, DeepFusion~\cite{li2022deepfusion}, FUTR3D~\cite{chen2023futr3d}, UniSeg~\cite{liu2023uniseg}) attempt to mitigate the problem of feature misalignment by a learnable sensor fusion, they still rely on calibration information to perform projection. In contrast, our \shortname 
% In contrast, our \shortname processes the multi-sensor information in implicit feature space and learns correspondence between different sensors by cross-attention layers.
% By contrast, our method fuses the multi-sensor information by learning the feature alignment implicitly under the guidance of 2D-3D reconstruction, which does not depend on calibration information and is therefore not subject to calibration errors. Although the image and point cloud contain a wealth of complementary information, there has been a lack of exploration in fusing multi-modal data for 3D semantic occupancy prediction.

\begin{figure*}[!ht]
\centering
    \includegraphics[width=0.97\linewidth]{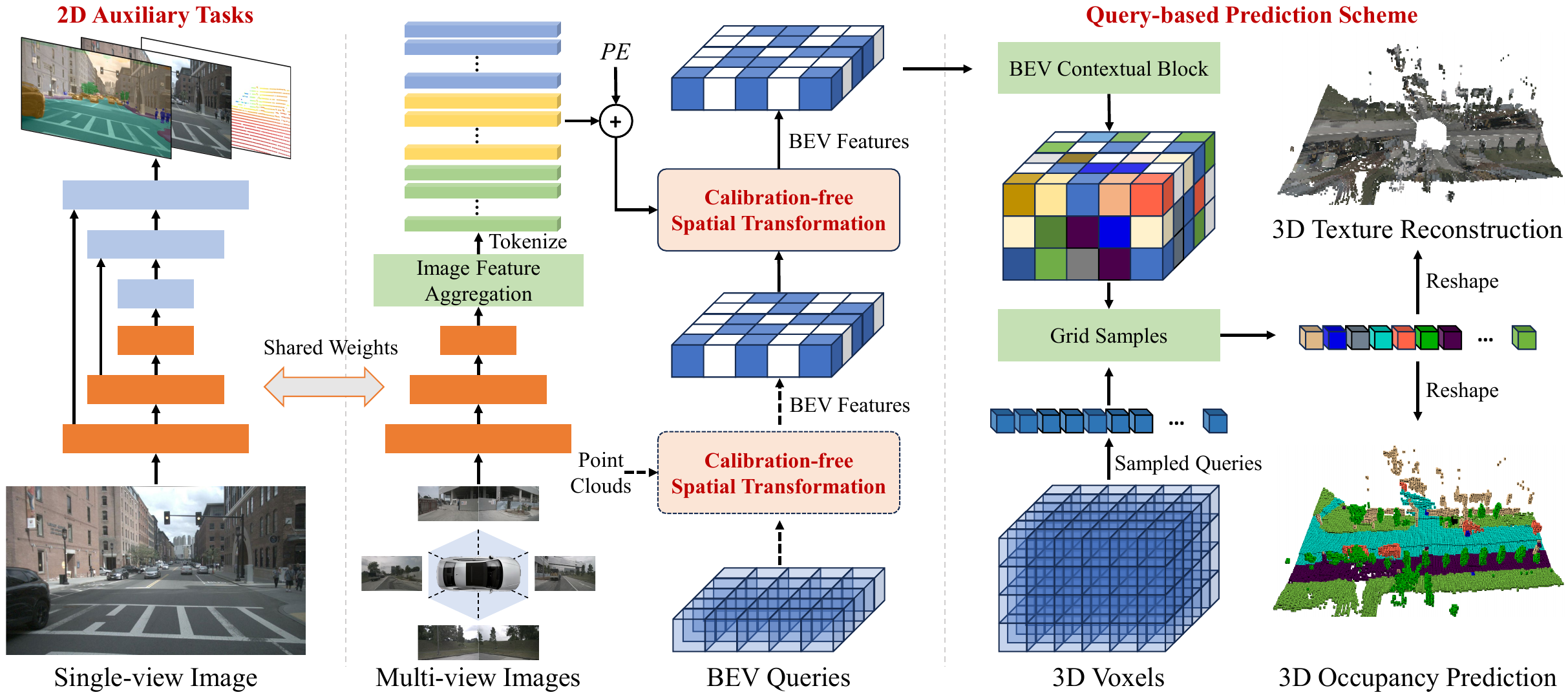}
    \caption{Overview of the proposed Robust and Efficient Occupancy (\shortname) Prediction. We first extract image features from multi-cameras using a pre-trained 2D encoder. Then, we aggregate the image features with a feature aggregation module. Both multi-cameras and LiDAR features are efficiently transformed to the BEV plane by the calibration-free spatial transformation modules. Third, we introduce 2D/3D auxiliary training tasks to ease the spatial projection and improve the model performance. Last, we use a query-based prediction scheme to efficiently generate predictions of queried voxels sampled from 3D space.}
    \label{fig:arch_overview}
\end{figure*}
% \noindent\textbf{Query-based \cst{3D} Perception}
% Query-based paradigm is one of the main effective ways to elevate 2D image information to 3D space. DETR3D~\cite{wang2022detr3d} uses a sparse set of 3D object queries to index into 2D image features via backward geometric projection and aggregates them. 

% \noindent\textbf{3D Semantic Scene Completion}

\noindent\textbf{2D-to-3D spatial transformation}
Vision-based perception has recently gained much attention due to its cost-effectiveness and rich appearance information. A crucial step in these methods is to project 2D image features into 3D space. One paradigm follows the forward projection. Lift-Splat-Shoot~\cite{philion2020lift} performs 2D-to-3D projection based on the estimated pixel-wise depth distribution. BEVDepth~\cite{li2023bevdepth} leverages depth supervision derived from point clouds to guide depth learning. BEVStereo~\cite{li2023bevstereo} further proposes a dynamic temporal stereo technique to enhance depth estimation. MatrixVT~\cite{zhou2023matrixvt} introduces Prime Extraction and the Ring \& Ray Decomposition to accelerate the feature transformation. Another paradigm follows the backward projection. OFT~\cite{roddick2018orthographic} and ImVoxelNet~\cite{rukhovich2022imvoxelnet} reconstruct 2D image features to 3D voxels along the camera rays with the corresponding intrinsic and extrinsic camera parameters. DETR3D~\cite{wang2022detr3d} projects the predefined learnable 3D queries onto 2D images to sample the corresponding features. BEVFormer~\cite{li2022bevformer} designs deformable cross-attention to aggregate the spatial features from multi-camera images and generate BEV features. FB-BEV~\cite{li2023fb} leverages the two paradigms and performs the forward-backward projection to transform the multi-view image features into BEV representation. In addition, PETR~\cite{liu2022petr} proposes to encode 3D coordinate information into 2D image features to generate 3D position-aware features. Note that existing works still require calibration information of cameras to perform 2D-to-3D projection, thus inevitably affected by the calibration noise. In contrast, Zhou~\etal\cite{zhou2022cross} implicitly model the spatial correspondence between multi-camera and map-view representation for 2D map-view semantic segmentation. We further exploit such an idea on complex multi-modal 3D semantic occupancy prediction.

\noindent\textbf{3D semantic occupancy prediction}
Recently, researchers have investigated 3D semantic occupancy prediction that provides a comprehensive and coherent perception of the outdoor environment~\cite{li2022bevformer,huang2023tri,zuo2023pointocc}. BEVFormer~\cite{li2022bevformer} first defines learnable grid-shaped BEV (bird's-eye-view) queries that capture spatial information from multi-view images with deformable attention layers. TPVFormer~\cite{huang2023tri} further designs a tri-perspective view (TPV) representation that accompanies BEV with two additional perpendicular planes to describe a stereoscopic 3D scene. SurroundOcc~\cite{wei2023surroundocc} defines voxel queries rather than BEV queries to fuse multi-camera information with 2D-to-3D spatial attention and applies 3D convolution to construct 3D volume features in a multi-scale fashion. VoxFormer~\cite{li2023voxformer} adopts a two-stage design where a query proposal network picks out reliable voxel queries and a sparse-to-dense MAE-like architecture generates dense semantic voxels. However, these methods rely on accurate sensor calibration information to compute the reference points for multi-view image feature aggregation. The model performance may degrade due to the calibration noise in real application scenarios. Unlike these works, our method directly aggregates the multi-view image features by vanilla cross-attention layers, discarding the reference points. To the best of our knowledge, our \shortname is the first work to exploit calibration-free multi-sensor feature fusion for 3D semantic occupancy prediction.

% However, their receptive field is limited because they rely on camera transformation matrices to link 3D query positions to multi-view images for reference points. Compared to the aforementioned works, our method performs global cross-attention on 3D queries and 2D image features, which provides a global perception. In addition, our method can easily aggregate information from other sensors.
% To the best of our knowledge, our \shortname is the first work to adapt the fully query-based scheme for 3D semantic occupancy prediction.

% ----------------------------------------------------
\section{Proposed Method}
As illustrated in Figure~\ref{fig:arch_overview}, our \shortname consists of a feature aggregation module (Section~\ref{sec:feature_agg}), calibration-free spatial transformation for both cameras and LiDAR(Section~\ref{sec:spatial_transform}), and query-based prediction scheme (Section~\ref{sec:coarse_to_fine}). Different from existing methods~\cite{cao2022monoscene,huang2023tri} that only use 3D semantic occupancy labels for supervision, we exploit 2D and 3D auxiliary training tasks including depth regression, semantic segmentation, and texture reconstruction (Section~\ref{sec:texture_guidance}) to help the image backbone capture spatial and semantic features. The overall training scheme of our method is shown in Algorithm~\ref{alg:proposed_method}.

\begin{algorithm}[t]
    \caption{General Scheme of \shortname}
    \begin{algorithmic}[1]
    \REQUIRE Training data $\{\mI, \bP, \bO\}$,  sampled queries $\bV^{fine}$, hyper parameter $\lambda_{focal},\lambda_{dice},\lambda_{rgb},\lambda_{depth}$ and model $\text{M}$.
    % \STATE \textit{// label preprocessing}
    % \STATE Project the occupied voxels to multi-view images to obtain 3D texture cues $\bO_{rgb}$. 
    % \STATE Project point clouds to the multi-view images to obtain the ground truth $\bY_{sem},\bY_{rgb},\bY_{depth}$ for 2D auxiliary tasks. 
    % \STATE \textit{// training}
    \WHILE{\textit{not convergent}}
        \STATE Sample an image $\bI_n$ from multi-view images $\mI$ and obtain the 2D predictions.
        \STATE Compute the objectives of 2D auxiliary tasks according to Eqs.(\ref{eq:2d_obj_sem}) and (\ref{eq:2d_obj_depth_rgb}).
        \STATE Compute the voxelized point clouds features $\bV^{LiDAR}$ and obtain the LiDAR feature token $\bT^{LiDAR}$ using Eq.(\ref{eq:lidar_token}).
        \STATE Project the LiDAR feature to BEV plane by the proposed \shortname module and obtain $\bB^{LiDAR}$.
        \STATE Aggregate the image features from pre-trained image backbones and obtain the image tokens $\bT^{cam}_n$ \wrt the multi-view images.
        \STATE Use BEV features \wrt LiDAR as queries and project the multi-view image features to BEV plane by \shortname module.
        \STATE Lift the BEV features to voxel space and compute coarse predictions by Eqs.(\ref{eq:bev_context}) and (\ref{eq:coarse_mlp}).
        \STATE Compute the fine-grained predictions by grid sample and compute the 3D predictions by Eqs.(\ref{eq:fine_sem_geo}) and (\ref{eq:fine_rgb}).
        \STATE Compute the objectives \wrt the 2D/3D tasks and optimize model $M$ by minimizing the objective in Eq.(\ref{eq:loss_total}).
    \ENDWHILE
    \end{algorithmic}
    \label{alg:proposed_method}
    % \vskip -0.1in
\end{algorithm}

\subsection{Problem definition}
Let $\mI=\{\bI_1, \bI_2, \dots, \bI_N\}$ be the input multi-view images, where $\bI_n\in\mmR^{3\times H_{in}\times W_{in}}$ denotes the RGB image \wrt the $n$-th camera. $H_{in}, W_{in}$ denote the height, and width of the image. Let $\bP\in\mmR^{N_{point}\times 4}$ be a point cloud from LiDAR with $N_{point}$ points.
The target of 3D semantic occupancy prediction is to predict the semantic class of each voxel in the $\bO\in \mmR^{H\times W\times D}$ voxel space using multi-sensor information (\eg, cameras and LiDAR) and indicate whether the voxel is empty or belongs to a specific semantic class $s\in\{0, 1, 2, 3,\dots, S\}$. Here, $S$ is the number of semantic classes, and $0$ often denotes the free voxel. $H,W,D$ denote the height, width, and length of the 3D voxel space, respectively. 

Note that it is unnecessary to obtain the full prediction results of 3D space in practice, one can only compute the occupancy results \wrt voxels in the interested region. Specifically, given voxels in 3D interested region, the occupancy prediction is computed by 
\begin{equation}
    \hat{\bO} = \rm{M}(\bV, \mI, \bP),
\end{equation}
where $\bV\in\mmR^{Q\times 3}$ denotes the voxels with position $(h,w,d)$ in the 3D space. $Q=H\times W\times D$ indicates the number of voxels. $\rm{M}(\cdot)$ denotes the deep model.

\subsection{Image feature aggregation}
\label{sec:feature_agg}
Following~\cite{huang2023tri}, we adopt a pre-trained image encoder (\eg, ResNet-50\cite{he2016deep}) to extract 2D image features. Different from existing methods~\cite{li2022bevformer}, we do not use a feature pyramid network (FPN) to aggregate the multi-level image features by considering the model efficiency. Instead, we use the features $\bH_n^l$ from the $l$-layer of the image backbone. $\bH_n\in\mmR^{C_l\times H_l\times W_l}$ indicate the features \wrt the $l$-layer and $n$-camera. Then, we apply the spatial attention module (SAM)~\cite{bochkovskiy2020yolov4} to augment the features and use a self-attention layer to learn the inner correspondence of features~\wrt each camera.

\subsection{Calibration-free spatial transformation}
\label{sec:spatial_transform}
\begin{figure}
\centering
    \includegraphics[width=1.0\linewidth]{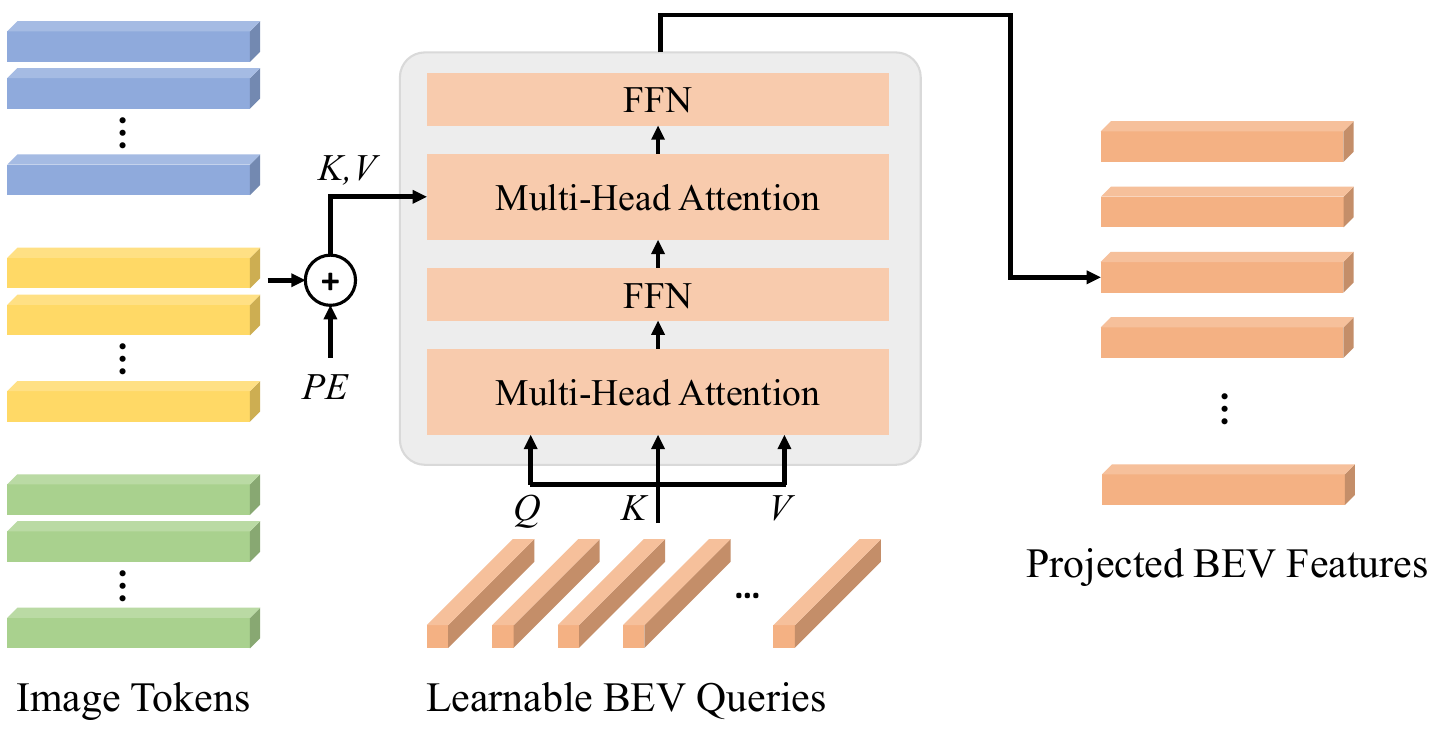}
    \caption{Illustration of \lowername for multi-cameras. \textit{PE} indicates the positional embeddings. The features from multi-view images are projected to a predefined BEV plane with the vanilla attention scheme. For multi-sensor fusion, the learnable BEV queries are replaced by the projected BEV LiDAR features.}
    \label{fig:spatial_cam}
\end{figure}
\noindent\textbf{Spatial transformation for multi-cameras.} To lift the 2D image features to 3D space, a common practice is to use deformable cross-attention (DCA)~\cite{li2022bevformer,huang2023tri} with reference points determined by the calibration information(\ie, camera extrinsic and intrinsic). As a result, the model performance is degraded when introducing noise to the sensor calibration. To address this issue, we proposed a calibration-free spatial transformation that built upon vanilla cross-attention layers to obtain 3D features without using sensor calibration (See Figure~\ref{fig:spatial_cam}). 

Let $\bF_n^{cam}\in\mmR^{C_f\times H_f \times W_f}$ be the merged image features~\wrt the $n$-th camera. $C_f$ indicates the number of channels of the image features. $H_f$ and $W_f$ denote the height and width of the features, respectively. 
Then, the image features are converted to a set of tokens by flattening the spatial dimensions. The feature tokens $\bT_n^{cam}\in \mmR^{N_t\times C_t}$~\wrt the $n$-th image features are computed by 
\begin{equation}
    \bT_n^{cam} = \text{Tokenizer}(\bF_n^{cam}),
    \label{eq:img_token}
\end{equation}
where $N_t$ and $C_t$ indicate the length and embedding dimensions of the feature tokens, respectively. 

% As demonstrated in Figure~\ref{fig:motivate_token_compress}, there are a lot of overlap areas between the multi-view images from different cameras. Even for an image from a single camera, there also exist similar features inside the image. Thus, it is possible to aggregate the features from different images by fixed-length feature tokens and improve the efficiency of the cross-attention layer.

Similar to BevFormer~\cite{li2022bevformer}, we adopt learnable embedding $\bQ\in\mmR^{N_{bev}\times C_t}$ as queries to aggregate the feature tokens from multi-cameras to a predefined compact BEV plane. $N_{bev}=H_{bev}\times W_{bev}$ denotes the length of BEV queries. Given points $\bP\in\mmR^{N_{bev}\times 2}$ from the BEV plane, we compute the BEV queries $\bQ$ by 
\begin{equation}
    \bQ = \text{PositionEncoding}(\bP)+\bW,
\end{equation}
where $\bW\in\mmR^{N_{bev}\times C_t}$ denotes the learnable positional embedding. $\text{PositionEncoding}(\cdot)$ indicate positional encoding layer~\cite{vaswani2017attention}.

With the vanilla cross-attention scheme, the projected image BEV feature $\bB^{cam}\in \mmR^{N_{bev} \times C_t}$ is computed by
\begin{equation}
    \bB^{cam} = \text{SpatialCrossAttn}(\bQ, \mT).
    \label{eq:cam_attn}
\end{equation}

% where $\bW\in\mmR^{N_{cam}\times C_t}$ indicates the learnable embedding. $N_{cam}$ indicates the length of learnable embedding.

\begin{figure}
\centering
    \includegraphics[width=1.0\linewidth]{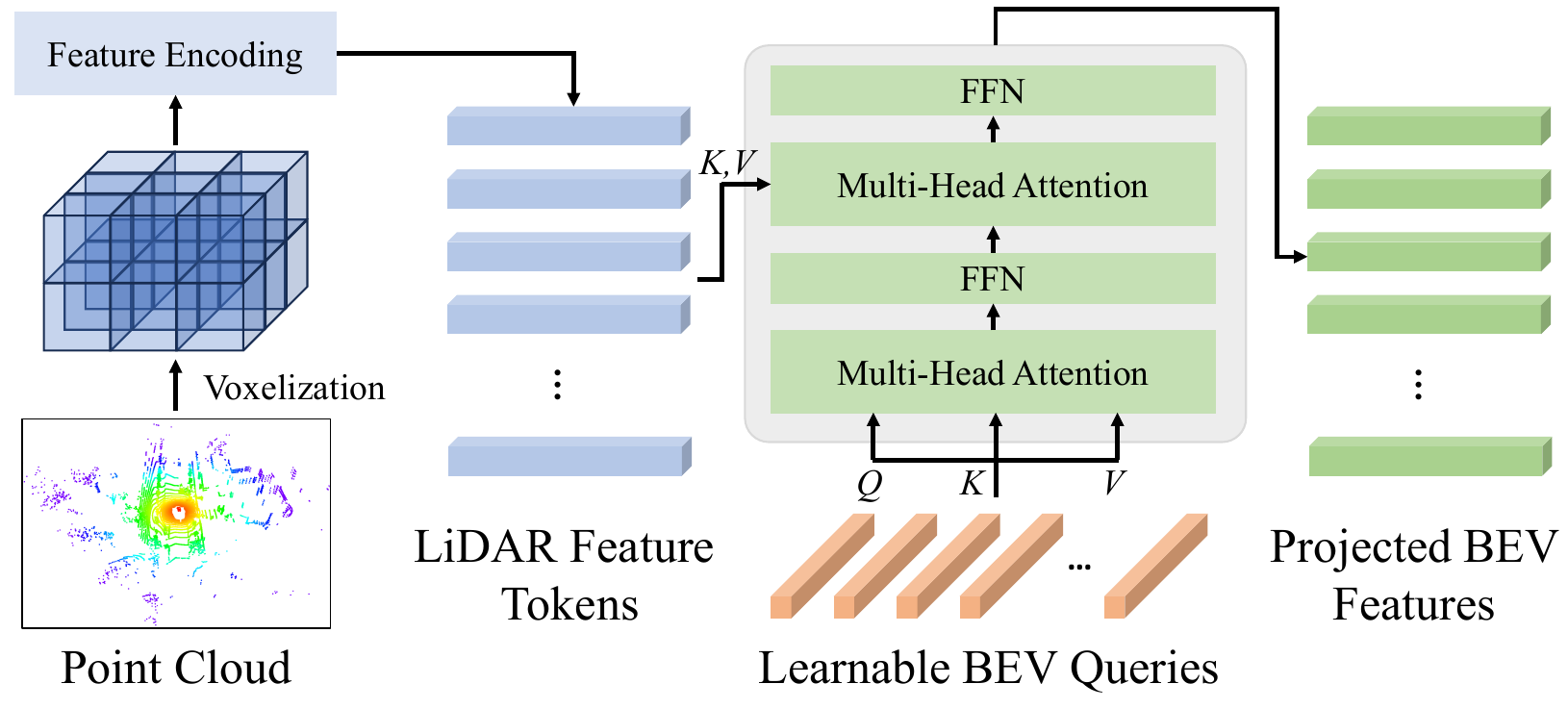}
    \caption{Illustration of \lowername for LiDAR. We assume that the LiDAR coordinate differs from the ground truth coordinate. Therefore, we use \shortname to project the LiDAR features to the predefined BEV plane with an attention scheme.}
    \label{fig:lidar_fusion}
\end{figure}
\noindent\textbf{Spatial transformation for LiDAR}
With the proposed calibration-free spatial transformation, we can obtain the BEV representations from multi-view 2D images. As point clouds provide complementary spatial information of images, one can further fuse the LiDAR features to improve the model performance. However, since the high-resolution LiDAR is expensive, we only consider fusing a small number of voxelized point clouds with lower resolution to reduce the dependency on LiDAR.\footnote{See experiments in Table~\ref{tab:effect_num_voxels}.}

Given a point cloud from LiDAR with $N_{point}$ points. Each point in the point cloud contains a 3D coordinate $(x, y, z)$ and an intensity value $(i)$. As shown in Figure~\ref{fig:lidar_fusion}, we voxelize the point clouds and randomly sample $N_{LiDAR}$ voxels. Each LiDAR voxel contains the coordinates $\{x,y,z,i\}$ of the point, where $i$ indicates the intensity value. We also compute the density and range value $r=\sqrt{x^2+y^2+z^2}$ of each voxel. Then, we use a position encoding layer~\cite{vaswani2017attention} to convert the voxels into feature tokens. The LiDAR feature tokens $\bT^{LiDAR}\in \mmR^{N_{LiDAR}\times C_t}$ is computed by 
\begin{equation}
    \bT^{LiDAR} = \text{PositionEncoding}(\bV^{LiDAR}),
    \label{eq:lidar_token}
\end{equation}
where $N_{LiDAR}$ indicates the length of LiDAR tokens. $\bV^{LiDAR}\in\mmR^{N_{LiDAR}\times 6}$ denotes a set of LiDAR voxels.

Note that we cannot compute the BEV representation of point clouds and fuse the image features by concatenation directly when the extrinsic of LiDAR is missing. Besides, the LiDAR coordinates may also differ from the ground truth coordinates. Similar to the CFL module of cameras, we also use vanilla attention modules to align the point cloud features to ground truth coordinates and conduct feature fusion with cameras. Specifically, for LiDAR-only implementation, we compute the BEV representation of LiDAR $\bB^{LiDAR}\in \mmR^{N_{bev}\times C_t}$ by 
\begin{equation}
    \bB^{LiDAR} = \text{SpatialCrossAttn}(\bQ, \bT^{LiDAR}).
\end{equation}

To multi-modal fusion, we use $\bB^{LiDAR}$ as BEV queries to lift the image feature to the BEV plane. Thus, similar to Eq.(\ref{eq:cam_attn}), the fused BEV features are computed by 
\begin{equation}
    \bB^{fuse} = \text{SpatialCrossAttn}(\bB^{LiDAR}, \mT).
\end{equation}

\subsection{Query-based prediction scheme} 
\label{sec:coarse_to_fine}
Due to the huge number of predicted voxels (\eg, $10^5$ voxels on Occ3D-nuScenes and $10^7$ voxels on OpenOpccuancy), it would be computationally expensive to predict fine-grain occupancy prediction directly. Besides, it is unnecessary to predict all voxels with fine grid size as the 3D space is extremely sparse. Therefore, we first introduce BEV context blocks to enhance the contextual feature of BEV representation and lift the BEV features to voxel representations by reshaping the feature channels. Then, we use the MLP layer to generate coarse predictions with low resolutions from the voxelized features. Last, we compute the fine-grained prediction of queried voxels from low-resolution coarse predictions by grid sample.

% As the 3D space is extremely sparse, it is unnecessary to predict all voxels with fine grid size. As shown in Table~\ref{tab:nus_empty_rate}, about 95.13\% of voxels are empty on nuScenes under a grid size of 0.4$m$. Similar to~\cite{tian2023occ3d}, we introduce a coarse-to-fine query scheme to reduce the number of queried voxels. First, we queried all voxels with a coarse grid size to obtain the geometry occupancy of voxels. Then, we generate the coarse occupied voxels to the fine one with a fine grid size and predict the semantics of the queried fine voxels. With the grid size of 0.4$m$ and 1.6$m$ for fine and coarse voxels, respectively, we can reduce the number of queried voxels from 640,000 to 178,448(10,000+2,632*64), which results in approximate 3.58$\times$ acceleration theoretically.

Let $\bB\in \mmR^{C_t \times H_{bev} \times W_{bev}}$ be the BEV feature reshaped from the outputs of spatial cross-attention. To enhance the contextual features of BEV representations, we adopt contextual blocks introduced in~\cite{cortinhal2020salsanext}. The enhance BEV feature $\bB^{contex}$ is computed by 
\begin{equation}
    \bB^{contex} = \text{ContextualBlock}(\bB).
    \label{eq:bev_context}
\end{equation}
where $\bB^{contex}\in\mmR^{C_{contex}\times H_{bev} \times W_{bev}}$. Then, we split the channels of $\bB^{contex}$ to obtain voxel representation $\bV^{coarse}\in \mmR^{C_{coarse}\times H_{bev}\times W_{bev} \times D_{coarse}}$. Here, $C_{contex}=C_{coarse}\times D_{coarse}$ denotes the number of channels~\wrt BEV features. $C_{coarse}$ denotes the channels of voxel features. 

Then, we adopt the MLP layer to obtain occupancy predictions. The coarse semantic prediction $\hat{\bO}^{coarse}_{sem}\in \mmR^{(S+1)\times H_{coarse}\times W_{coarse} \times D_{coarse}}$ is computed by 
\begin{equation}
    \hat{\bO}^{coarse}_{sem} = \text{MLP}(\bV^{coarse}).
    \label{eq:coarse_mlp}
\end{equation}
Alternatively, we also obtain the geometry prediction $\hat{\bO}^{coarse}_{geo}\in \mmR^{2\times H_{coarse}\times W_{coarse} \times D_{coarse}}$ from the semantic predictions. 

Given $\bV^{fine}\in\mmR^{N_{fine}\times 3}$ sampled from the fine-grained voxel space, we further interpolate the fine predictions with a grid sample. Specifically, the geometry and semantic fine-grained predictions are computed by 
\begin{equation}
    \left\{
    \begin{aligned}
        & \hat{\bO}^{fine}_{sem} = \text{GridSample}(\hat{\bO}^{coarse}_{sem}, \bV^{fine}), \\
        & \hat{\bO}^{fine}_{geo} = \text{GridSample}(\hat{\bO}^{coarse}_{geo}, \bV^{fine}).
    \end{aligned}
    \right.
    \label{eq:fine_sem_geo}
\end{equation}

\subsection{Construction of 2D/3D auxiliary training tasks}
In semantic occupancy prediction, it is challenging to learn both the 2D-to-3D projection and semantics from multi-view images without spatial prior. In this section, we first propose a 3D texture reconstruction task to ease the learning of 2D-to-3D projection. Then, we introduce the 2D auxiliary training tasks to enhance both the semantic and spatial features of the 2D image backbone. 

\noindent\textbf{3D texture reconstruction}
\label{sec:texture_guidance}
Motivated by the scheme of auto-encoder~\cite{he2022masked}, we model the learning of spatial correspondence by 2D-to-3D texture reconstruction. Similar to Eqs.(\ref{eq:coarse_mlp}) and (\ref{eq:fine_sem_geo}), we predict the texture of voxels by
\begin{equation}
    \hat{\bO}_{rgb} = \text{GridSample}(\text{MLP}(\bV^{coarse}), \bV^{fine}),
    \label{eq:fine_rgb}
\end{equation}
where $\hat{\bO}_{rgb}\in \mmR^{N_{fine}\times 3}$ indicates the predicted RGB color~\wrt the voxels $\bV^{fine}$. Let $\bO_{rgb}$ be the ground truth color of the queried voxel (\ie, 3D texture cues). The auxiliary loss of the 3D texture reconstruction task is defined as
\begin{equation}
    \mL^{3D}_{rgb} = \text{SmoothL1}(\hat{\bO}_{rgb},\bO_{rgb}).
    \label{eq:loss_rgb}
\end{equation}

Note that the 3D texture cues are from the observations of the physical world and can be obtained by scene reconstruction~\cite{lu2023large}. For convenience, we project the image to 3D voxels to construct the 3D texture cues in this work.

\noindent\textbf{2D auxiliary training tasks.}
By considering the model efficiency, we do not adopt a heavy 3D decoder that is computationally expensive for occupancy predictions. However, it may also result in unsatisfied model performance. To address this issue, we propose three 2D auxiliary tasks to enhance the discrimination power of 2D backbone, including semantic segmentation, depth regression, and texture reconstruction.

Given $\bI_n\in\mmR^{3\times H_{in}\times W_{in}}$ sampled from multi-view images $\mI$, we adopt a FPN to fuse the multi-level features $\mH_n=\{\bH_n^l\}_{l=1}^{L}$ from the image backbone obtain the 2D predictions. For the semantic segmentation task, we use multi-class focal loss and dice loss as objectives. Thus, the objectives~\wrt the semantic segmentation task are formulated as
\begin{equation}
\begin{aligned}
    \mL_{sem}^{2D} =&\lambda_{focal} \mL_{focal}(\hat{\bY}_{sem}, \bY_{sem})\\
    & + \lambda_{dice} \mL_{dice}(\hat{\bY}_{sem}, \bY_{sem})),
\end{aligned}
\label{eq:2d_obj_sem}
\end{equation}
where $\hat{\bY}_{sem}\in \mmR^{S\times H_{in}\times W_{in}}$ and  $\bY_{sem}\in \mmR^{S\times H_{in}\times W_{in}}$ denote the predictions and ground truth of 2D semantic segmentation. $\lambda_{focal}$ and $\lambda_{dice}$ are the hyper parameters to balance the two terms. Following~\cite{zhuang2021perception}, we project the LiDAR semantic labels to the image to obtain the 2D semantic labels.

For both the depth regression and texture reconstruction tasks, we use the smooth L1 loss as the objective. Therefore, the object functions are formulated as
\begin{equation}
    \left\{
    \begin{aligned}
        & \mL_{depth}^{2D} = \lambda_{depth} \text{SmoothL1}(\hat{\bY}_{depth}, \bY_{depth}), \\
        & \mL_{rgb}^{2D} = \lambda_{rgb} \text{SmoothL1}(\hat{\bY}_{rgb}, \bY_{rgb}).
    \end{aligned}
    \right.
    \label{eq:2d_obj_depth_rgb}
\end{equation}
where $\hat{\bY}_{depth}$ and $\bY_{depth}$ denote the predictions and labels of depth regression, respectively. $\hat{\bY}_{rgb}$ and $\bY_{rgb}$ denote the predictions and labels of texture reconstruction, respectively. $\lambda_{depth}$ and $\lambda_{rgb}$ are the hyper parameters to balance the objectives. $\bY_{depth}$ is obtained by projecting the point clouds to 2D images. The decoder \wrt 2D auxiliary training is dropped to reduce model complexity during inference.

\subsection{Formulation of objective}
To optimize the occupancy network, we use focal loss~\cite{lin2017focal} and dice loss~\cite{sudre2017generalised} as objectives. The objective of semantic occupancy prediction is defined as 
\begin{equation}
    \begin{split}
        \mL^{3D}_{sem} =& \lambda_{focal}\mL_{wce}(\hat{\bO}^{fine}_{sem},\bO^{fine}_{sem}) \\
        &+ \lambda_{dice} \mL_{dice}(\hat{\bO}^{fine}_{sem},\bO^{fine}_{sem}).
    \end{split}
\end{equation}
Similar to the semantic occupancy predictions, the objective of geometry predictions is defined as
\begin{equation}
    \begin{split}
        \mL^{3D}_{geo} =& \lambda_{focal}\mL_{wce}(\hat{\bO}^{fine}_{geo},\bO^{fine}_{geo}) \\
        &+ \lambda_{dice} \mL_{dice}(\hat{\bO}^{fine}_{geo},\bO^{fine}_{geo}).
    \end{split}
\end{equation}

Considering the objective of all the above tasks, we formulate the object function of our proposed \shortname as 
\begin{equation}
    \mL=\mL^{3D}_{geo} + \mL^{3D}_{sem} +\mL^{3D}_{rgb}+\mL^{2D}_{sem}+\mL^{2D}_{depth}+\mL^{2D}_{rgb}.
    \label{eq:loss_total}
\end{equation}

\section{Experiments}
\subsection{Data sets and evaluation metrics}
\begin{table}
    \centering
    \caption{Comparisons of three benchmark data sets.}
    \scalebox{0.85}{
    \begin{tabular}{c|ccc}
    \hline
    Data set   & OpenOccupancy & Occ3D-nuScenes & SemanticKITTI\\
    \hline
    Grid size & 0.2$m$ & 0.4$m$ & 0.2$m$ \\
    \#Point Clouds  & 34720  & 34720 & 122820\\
    \#Voxelized Point Clouds & 11088 & 6811 & 27140 \\
    Compression Rate & 31.9\% & 19.6\% & 22.1\% \\
    \hline
    \end{tabular}
    }
    \label{tab:cmp_dataset}
\end{table}
\noindent\textbf{Data sets}
Following~\cite{huang2023tri,tian2023occ3d,wang2023openoccupancy}, we evaluate our method on OpenOccupancy~\cite{wang2023openoccupancy}, Occ3D-nuScenes~\cite{tian2023occ3d}, and SemanticKITTI Scene Completion~\cite{behley2019semantickitti}. 

\textbf{OpenOccupancy} is an occupancy prediction data set based on nuScenes, which provides 360-degree point clouds and images. The dataset has a scene range from [-51.2$m$, -51.2$m$, -5$m$] to [51.2$m$, 51.2$m$, 3$m$]. The 3D space is voxelized to $512\times 512 \times 40$ with a grid size of $0.2m$. It contains 700 scenes for training and 150 scenes for validation, respectively. Besides, it has 16 classes for semantic occupancy prediction.
\textbf{Occ3D-nuScenes} is a 3D semantic occupancy prediction dataset based on nuScenes~\cite{caesar2020nuscenes}, but with a scene range from [-40$m$, -40$m$, -1$m$] to [40$m$, 40$m$, 5.4$m$] along X, Y, and Z axes, which is voxelized into 200×200×16 given a voxel size of 0.4$m$. Note that Occ3D-nuScenes has 16 semantic classes and an additional unknown object class. Therefore, it has 17 semantic classes. 
\textbf{SemanticKITTI} is a large-scale outdoor data set based on the KITTI Odometry Benchmark~\cite{geiger2012we}, which provides 360-degree point clouds and front-view images.  For the semantic scene completion task~\cite{behley2019semantickitti}, the dataset has a valid evaluation range from [0$m$, -25.6$m$, -2$m$] to [51.2$m$, 25.6$m$, 4.4$m$] along X, Y, and Z axes, which is voxelized into 256×256×32. Each voxel has a size of 0.2m×0.2m×0.2m and is assigned a label from 21 classes (19 semantics, 1 empty, 1 unknown). SemanticKITTI divides its 22 sequences into three sets: sequences 00-10 except 08 for training, sequence 08 for validation, and sequences 11-21 for testing, respectively.

\noindent\textbf{Evaluation Metrics}
Following~\cite{cao2022monoscene,huang2023tri}, we use the intersection over union of scene completion (SC IoU) that ignores the semantic labels to evaluate the quality of scene completion on SemanticKITTI. Besides, we also report the semantic scene completion mean IoU (SSC mIoU) of all semantic classes on SemanticKITTI. For Occ3D-nuScenes, we report the mean IoU (mIoU) of 16 semantic classes~\cite{tian2023occ3d}. For OpenOccupancy, we follow the common practice of ~\cite{wang2023openoccupancy} and report both the mean IoU(mIoU) of 17 semantic classes and the IoU of geometry occupancy prediction.

\subsection{Implementation details}
We implement our method using PyTorch\cite{paszke2019pytorch} and use ResNet-50~\cite{he2016deep} as the backbone of the 2D image encoder. The parameters of the image encoder are initialized by ImageNet pre-trained weights from~\cite{paszke2019pytorch}. The length of embeddings $C_t$ is set to 384 empirically. We train the networks using AdamW~\cite{loshchilov2017decoupled} for 25 epochs on all the benchmark data sets. The learning rate starts at 0.0004 and decays to 0 with a cosine policy~\cite{loshchilov2016sgdr}. The weight decay is 0.01. The batch size is 16 for nuScenes and 8 for SemanticKITTI, respectively. The grid sizes of BEV queries are set to 1.6$m$ and 0.8$m$ on nuScenes and SemanticKITTI, respectively. On all the data sets, $\lambda_{focal},\lambda_{dice}$ and $\lambda_{rgb}$ are set to 1.0, and $\lambda_{depth}$ is set to 2.0. The input images are resized to $448\times 800$ on nuScenes and $358\times 1184$ on SemanticKITTI, respectively. As shown in Table~\ref{tab:cmp_dataset}, we count the average number of points and voxels of the point clouds on three data sets. Therefore, we set the number of sampled voxelized point clouds to 10240, 5120, and 20480 on OpenOccupancy, Occ3D-nuScenes, and SemanticKITTI, respectively.

%-----------------------------------

\begin{table*}
\centering
\caption{Comparisons on OpenOccupancy validation set. Cons. Veh represents construction vehicle and Dri. Sur denotes driveable surface, respectively. The \textbf{bold} numbers indicate the best results. $\dagger$ indicates the results with multi-sweep point clouds.} 
% \vskip 0.15in
 % scalebox{0.6}
% \renewcommand\arraystretch{1.1}
  \scalebox{0.9}{
\begin{tabular}{l|c|cc|cccccccccccccccc}
\hline
Method 
& \rotatebox{90}{Modality} 
& \rotatebox{90}{IoU (\%)} 
& \rotatebox{90}{mIoU (\%)} 
& \rotatebox{90}{barrier} 
& \rotatebox{90}{bicycle} 
& \rotatebox{90}{bus} 
& \rotatebox{90}{car} 
& \rotatebox{90}{Cons. Veh} 
& \rotatebox{90}{motorcycle} 
& \rotatebox{90}{pedestrian} 
& \rotatebox{90}{traffic cone} 
& \rotatebox{90}{trailer} 
& \rotatebox{90}{truck} 
& \rotatebox{90}{Dri. Sur} 
& \rotatebox{90}{other flat} 
& \rotatebox{90}{sidewalk} 
& \rotatebox{90}{terrain} 
& \rotatebox{90}{manmade} 
& \rotatebox{90}{vegetation} 
\\ 

\hline\hline
MonoScene~\cite{cao2022monoscene} & C & 17.1 & 7.2 & 7.3 & 4.3 & 9.6 & 7.1 & 6.2 & 3.5 & 5.9 & 4.7 & 5.6 & 4.9 & 15.6 & 6.8 & 7.9 & 7.6 & 10.5 & 7.9 \\
TPVFormer~\cite{huang2023tri} & C & 15.1 & 8.3 & 9.7 & 4.5 & 11.5 & 10.7 & 5.5 & 4.6 & 6.3 & 5.4 & 6.9 & 6.9 & 14.1 & 9.8 & 8.9 & 9.0 & 9.9 & 8.5 \\
LMSCNet$^\dagger$~\cite{roldao2020lmscnet} & L & 26.7 & 11.8 & 12.9 & 5.2 & 12.8 & 12.6 & 6.6 & 4.9 & 6.3 & 6.5 & 8.8 & 7.7 & 24.3 & 12.7 & 16.5 & 14.5 & 14.2 & 22.1 \\
JS3C-Net$^\dagger$~\cite{yan2021sparse} & L & 29.6 & 12.7 & 14.5 & 4.4 & 13.5 & 12.0 & 7.8 & 4.4 & 7.3 & 6.9 & 9.2 & 9.2 & 27.4 & 15.8 & 15.9 & 16.4 & 14.0 & 24.8 \\
SparseOcc~\cite{tang2024sparseocc} & C & 21.8 & 14.1 & 16.1 & 9.3 & 15.1 & 18.6 & 7.3 & 9.4 & 11.2 & 9.4 & 7.2 & 13.0 & 31.8 & 21.7 & 20.7 & 18.8 & 6.1 & 10.6 \\
% RangeOcc & L & 30.1 & 19.0 & 18.8 & 10.5 & 19.2 & 21.7 & 10.5 & 17.0 & 19.7 & 11.0 & 13.0 & 17.7 & 32.6 & 21.4 & 22.3 & 22.0 & 20.7 & 25.4 \\
C-CONet~\cite{wang2023openoccupancy} & C & 21.6 & 13.6 & 13.6 & 8.4 & 14.7 & 18.3 & 7.1 & 11.0 & 11.8 & 8.8 & 5.2 & 13.0 & 32.7 & 21.1 & 20.1 & 17.6 & 5.1 & 8.4 \\
L-CONet$^\dagger$~\cite{wang2023openoccupancy} & L & 30.1 & 15.9 & 18.0 & 3.9 & 14.2 & 18.7 & 8.3 & 6.3 & 11.0 & 5.8 & 14.1 & 14.3 & 35.3 & 20.2 & 21.5 & 20.9 & 19.2 & 23.0 \\ 
M-CONet$^\dagger$~\cite{wang2023openoccupancy} & C\&L & 26.5 & 20.5 & 23.3 & 16.1 & 22.2 & 24.6 & 13.3 & 20.1 & 21.2 & \textbf{14.4} & 17.0 & 21.3 & 31.8 & 22.0 & 21.8 & 20.5 & 17.7 & 20.4 \\
C-OccGen~\cite{wang2024occgen} & C & 23.4 & 14.5 & 15.5 & 9.1 & 15.3 & 19.2 & 7.3 & 11.3 & 11.8 & 8.9 & 5.9 & 13.7 & 34.8 & 22.0 & 21.8 & 19.5 & 6.0 & 9.9 \\
L-OccGen$^\dagger$~\cite{wang2024occgen} & L & 31.6 & 16.8 & 18.8 & 5.1 & 14.8 & 19.6 & 7.0 & 7.7 & 11.5 & 6.7 & 13.9 & 14.6 & 36.4 & 22.1 & 22.8 & 22.3 & 20.6 & 24.5 \\
M-OccGen$^\dagger$~\cite{wang2024occgen} & C\&L & 30.3 & \textbf{22.0} & 24.9 & 16.4 & \textbf{22.5} & 26.1 & \textbf{14.0} & 20.1 & 21.6 & 14.6 & \textbf{17.4} & \textbf{21.9} & 35.8 & 24.5 & 24.7 & 24.0 & 20.5 & 23.5 \\
Co-Occ$^\dagger$~\cite{pan2024co} & C\&L & 30.6 & 21.9 & 26.5 & \textbf{16.8} & 22.3 & \textbf{27.0} & 10.1 & \textbf{20.9} & 20.7 & 14.5 & 16.4 & 21.6 & 36.9 & 23.5 & 25.5 & 23.7 & 20.5 & 23.5 \\
% PointOcc~\cite{zuo2023pointocc} & L & 30.9 & 20.5 & 22.4 & 12.9 & 19.9 & 24.5 & 10.2 & 20.2 & 27.0 & 12.6 & 13.8 & 19.2 & 32.4 & 19.5 & 22.4 & 21.9 & 22.5 & 26.7 \\
% PointOcc$^\dagger$~\cite{zuo2023pointocc} & L & 34.1 & 23.9 & 24.9 & 19.0 & 20.9 & 25.7 & 13.4 & 25.6 & 30.6 & 17.9 & 16.7 & 21.2 & 36.5 & 25.6 & 25.7 & 24.9 & 24.8 & 29.0 \\
\hline
\shortname(Ours) & C\&L & 31.7 & 21.3 & \textbf{24.9} & 13.1 & 22.0 & 25.0 & 11.5 & 19.7 & \textbf{22.4} & 11.1 & 16.2 & 21.5 & 35.2 & 22.9 & 24.5 & 23.9 & 21.9 & 25.5
\\
\shortname$^\dagger$(Ours) & C\&L & \textbf{33.7} & 21.2 & 24.4 & 11.1 & 20.9 & 24.6 & 11.2 & 16.3 & 17.5 & 11.4 & 16.0 & 20.8 & \textbf{38.5} & \textbf{25.5} & \textbf{26.0} & \textbf{25.9} & \textbf{22.6} & \textbf{27.1} \\
\hline
\end{tabular}
}
\label{tab:nus_openocc}
\end{table*}

\begin{table*}
\centering
\caption{3D semantic occupancy prediction results on Occ3D-nuScenes validation set. Cons. Veh represents construction vehicle and Dri. Sur denotes driveable surface, respectively. The \textbf{bold} numbers indicate the best results.} 
% \vskip 0.15in
 % scalebox{0.6}
% \renewcommand\arraystretch{1.1}
  \scalebox{0.83}{
\begin{tabular}{l|c|ccccccccccccccccc}
\hline
Method 
& \rotatebox{90}{mIoU (\%)} 
& \rotatebox{90}{others} 
& \rotatebox{90}{barrier} 
& \rotatebox{90}{bicycle} 
& \rotatebox{90}{bus} 
& \rotatebox{90}{car} 
& \rotatebox{90}{Cons. Veh} 
& \rotatebox{90}{motorcycle} 
& \rotatebox{90}{pedestrian} 
& \rotatebox{90}{traffic cone} 
& \rotatebox{90}{trailer} 
& \rotatebox{90}{truck} 
& \rotatebox{90}{Dri. Sur} 
& \rotatebox{90}{other flat} 
& \rotatebox{90}{sidewalk} 
& \rotatebox{90}{terrain} 
& \rotatebox{90}{manmade} 
& \rotatebox{90}{vegetation} 
\\ 

\hline\hline

MonoScene~\cite{cao2022monoscene} & 6.06 & 1.75 & 7.23 & 4.26 & 4.93 & 9.38 & 5.67 & 3.98 & 3.01 & 5.90 & 4.45 & 7.17 & 14.91 & 6.32 & 7.92 & 7.43 & 1.01 & 7.65 \\

BEVDet~\cite{huang2021bevdet} & 19.38 & 4.39 & 30.31 & 0.23 & 32.26 & 34.47 & 12.97 & 10.34 & 10.36 & 6.26 & 8.93 & 23.65 & 52.27 & 24.61 & 26.06 & 22.31 & 15.04 & 15.10 \\

OccFormer~\cite{zhang2023occformer} & 21.93 & 5.94 & 30.29 & 12.32 & 34.40 & 39.17 & 14.44 & 16.45 & 17.22 & 9.27 & 13.90 & 26.36 & 50.99 & 30.96 & 34.66 & 22.73 & 6.76 & 6.97 \\

BEVFormer~\cite{li2022bevformer} & 26.88 & 5.85 & 37.83 & 17.87 & 40.44 & 42.43 & 7.36 & 23.88 & 21.81 & 20.98 & 22.38 & 30.70 & 55.35 & 28.36 & 36.00  &28.06 & 20.04 & 17.69 \\

TPVFormer~\cite{huang2023tri} & 27.83 & 7.22 & 38.90 & 13.67 & 40.78 & 45.90 & 17.23 & 19.99 & 18.85 & 14.30 & 26.69 & 34.17 & 55.65 & 35.47 & 37.55 & 30.70 & 19.40 & 16.78 \\

CTF-Occ~\cite{tian2023occ3d} & 28.53 & 8.09 & 39.33 & 20.56 & 38.29 & 42.24 & 16.93 & 24.52 & 22.72 & 21.05 & 22.98 & 31.11 & 53.33 & 33.84 & 37.98 & 33.23 & 20.79 & 18.00 \\

FB-Occ~\cite{li2023fb} & 42.06 & \textbf{14.30} & 49.71 & \textbf{30.00} & 46.62 & 51.54 & 29.30 & 29.13 & 29.35 & 30.48 & 34.97 & 39.36 & 83.07 & \textbf{47.16} & 55.62 & 59.88 & 44.89 & 39.58 \\

PanoOcc~\cite{wang2024panoocc} & 42.13 & 11.67 & 50.48 & 29.64 & 49.44 & 55.52 & 23.29 & 33.26 & 30.55 & 30.99 & 34.43 & 42.57 & 83.31 & 44.23 & 54.40 & 56.04 & 45.94 & 40.40 \\

COTR~\cite{ma2024cotr} & 44.45 & 13.29 & 52.11 & 31.95 & 46.03 & 55.63 & \textbf{32.57} & 32.78 & 30.35 & \textbf{34.09} & 37.72 & 41.84 & \textbf{84.48} & 46.19 & \textbf{57.55} & \textbf{60.67} & 51.99 & 46.33 \\
\hline

\shortname(Ours) & \textbf{46.51} & 12.39 & \textbf{55.69} & 24.10 & \textbf{56.12} & \textbf{57.55} & 32.20 & \textbf{35.26} & \textbf{40.68} & 24.95 & \textbf{47.16} & \textbf{49.07} & 81.45 & 43.35 & 52.17 & 56.87 & \textbf{61.89} & \textbf{59.85} 
\\
\hline
\end{tabular}
}
\label{tab:occ3d_nuscenes_results}
\end{table*}

\subsection{Comparisons on benchmark data sets}
\noindent\textbf{Results on OpenOccupancy.}
To compare our method with the state-of-the-art methods (~\eg, M-CONet~\cite{wang2023openoccupancy}, M-OccGen~\cite{wang2024occgen}, and Co-Occ~\cite{pan2024co}), we evaluate our method on OpenOccuapncy~\cite{wang2023openoccupancy}. As shown in Table~\ref{tab:nus_openocc}, we report both the semantic metric mIoU and geometry metric IoU on OpenOccupancy. From the results, our \shortname outperforms existing fusion-based methods in geometry IoU and achieves comparable performance in semantic mIoU. Note that existing LiDAR-only and fusion-based methods often stack multi-sweep point clouds as inputs. For a fair comparison, we also report the performance of \shortname with 10-sweeps point clouds, and the number of sampled point cloud voxels is set to 40960. In this case, our \shortname outperforms the state-of-the-art fusion-based method (\ie, Co-Occ) by 2.8\% in geometry IoU.

\noindent\textbf{Results on Occ3D-nuScenes.}
Different from OpenOccupancy, Occ3d-nuScenes is a data set typically for camera-only methods as it considers the visible mask of multi-cameras. We also compare our \shortname with the advanced camera-centric methods on the Occ3D-nuScenes validation set. As shown in Table~\ref{tab:occ3d_nuscenes_results}, our \shortname achieves the best performance on Occ3D-nuScenes. Specifically, \shortname outperforms the state-of-the-art method, \ie, COTR~\cite{ma2024cotr}, by 2.06\% in mIoU. Note that the point clouds on nuScenes are collected by a 32-beam LiDAR and are very sparse. As we only fuse 5,120 voxels from point clouds, these results implicit that our \shortname can effectively reduce the dependency on high-resolution LiDAR.

\begin{table*}[htp]
\centering
\caption{Semantic scene completion results on SemanticKITTI validation set. * represents the RGB-inferred versions of these methods, which are reported in MonoScene~\cite{cao2022monoscene}. $\dagger$ denotes the reproduced result from TPVFormer~\cite{huang2023tri}. The \textbf{bold} numbers indicate the best results.} 
% \vskip 0.15in
 % scalebox{0.6}
% \renewcommand\arraystretch{1.1}
  \scalebox{0.76}{
\begin{tabular}{l|cc|ccccccccccccccccccc}
\hline
Method 
& \rotatebox{90}{SC IoU (\%)} 
& \rotatebox{90}{SSC mIoU (\%)} 
& \rotatebox{90}{road} 
& \rotatebox{90}{sidewalk} 
& \rotatebox{90}{parking} 
& \rotatebox{90}{other-ground} 
& \rotatebox{90}{building} 
& \rotatebox{90}{car} 
& \rotatebox{90}{truck} 
& \rotatebox{90}{bicycle} 
& \rotatebox{90}{motorcycle} 
& \rotatebox{90}{other-vehicle} 
& \rotatebox{90}{vegetation} 
& \rotatebox{90}{trunk} 
& \rotatebox{90}{terrain} 
& \rotatebox{90}{person} 
& \rotatebox{90}{bicyclist} 
& \rotatebox{90}{motorcyclist} 
& \rotatebox{90}{fence} 
& \rotatebox{90}{pole} 
& \rotatebox{90}{traffic-sign} 
\\ 

\hline\hline

LMSCNet*~\cite{roldao2020lmscnet} & 28.61 & 6.70 & 40.68 & 18.22 & 4.38 & 0.00 & 10.31 & 18.33 & 0.00 & 0.00 & 0.00 & 0.00 & 13.66 & 0.02 & 20.54 & 0.00 & 0.00 & 0.00 & 1.21 & 0.00 & 0.00  \\

3DSketch*~\cite{chen20203d} & 33.30 & 7.50 & 41.32 & 21.63 & 0.00 & 0.00 & 14.81 & 18.59 & 0.00 & 0.00 & 0.00 & 0.00 & 19.09 & 0.00 & 26.40 & 0.00 & 0.00 & 0.00 & 0.73 & 0.00 & 0.00 \\

AICNet*~\cite{li2020anisotropic} & 29.59 & 8.31 & 43.55 & 20.55 & 11.97 & 0.07 & 12.94 & 14.71 & 4.53 & 0.00 & 0.00 & 0.00 & 15.37 & 2.90 & 28.71 & 0.00 & 0.00 & 0.00 & 2.52 & 0.06 & 0.00 \\

JS3C-Net*~\cite{yan2021sparse} & 38.98 & 10.31 & 50.49 & 23.74 & 11.94 & 0.07 & 15.03 & 24.65 & 4.41 & 0.00 & 0.00 & 6.15 & 18.11 & 4.33 & 26.86 & 0.67 & 0.27 & 0.00 & 3.94 & 3.77 & 1.45 \\

MonoScene$^\dagger$~\cite{cao2022monoscene} & 36.86 & 11.08 & 56.52 & 26.72 & 14.27 & 0.46 & 14.09 & 23.26 & 6.98 & 0.61 & 0.45 & 1.48 & 17.89 & 2.81 & 29.64 & 1.86 & 1.20 & 0.00 & 5.84 & 4.14 & 2.25 \\

TPVFormer~\cite{huang2023tri} & 35.61 & 11.36 & 56.50 & 25.87 & 20.60 & 0.85 & 13.88 & 23.81 & 8.08 & 0.36 & 0.05 & 4.35 & 16.92 & 2.26 & 30.38 & 0.51 & 0.89 & 0.00 & 5.94 & 3.14 & 1.52 \\

VoxFormer-S~\cite{li2023voxformer} & 44.02 & 12.35 & 54.76 & 26.35 & 15.50 & 0.70 & 17.65 & 25.79 & 5.63 & 0.59 & 0.51 & 3.77 & 24.39 & 5.08 & 29.96 & 1.78 & 3.32 & 0.00 & 7.64 & 7.11 & 4.18 \\

VoxFormer-T~\cite{li2023voxformer} & 44.15 & 13.35 & 53.57 & 26.52 & 19.69 & 0.42 & 19.54 & 26.54 & 7.26 & 1.28 & 0.56 & 7.81 & 26.10 & 6.10 & 33.06 & 1.93 & 1.97 & 0.00 & 7.31 & 9.15 & 4.94 \\

SparseOcc~\cite{tang2024sparseocc} & 36.48 & 13.12 & 59.59 & 29.68 & 20.44 & 0.47 & 15.41 & 24.03 & 18.07 & 0.78 & 0.89 & 8.94 & 18.89 & 3.46 & 31.06 & \textbf{3.68} & 0.62 & 0.00 & 6.73 & 3.89 & 2.60 \\

OccFormer~\cite{zhang2023occformer} & 36.50 & 13.46 & 58.85 & 26.88 & 19.61 & 0.31 & 14.40 & 25.09 & \textbf{25.53} & 0.81 & 1.19 & 8.52 & 19.63 & 3.93 & 32.62 & 2.78 & 2.82 & 0.00 & 5.61 & 4.26 & 2.86 \\

HASSC-S~\cite{wang2024not} & 44.82 & 13.48 & 57.05 & 28.25 & 15.90 & 1.04 & 19.05 & 27.23 & 9.91 & 0.92 & 0.86 & 5.61 & 25.48 & 6.15 & 32.94 & 2.80 & \textbf{4.71} & 0.00 & 6.58 & 7.68 & 4.05 \\

H2GFormer-S~\cite{wang2024h2gformer} & 44.57 & 13.73 & 56.08 & 29.12 & 17.83 & 0.45 & 19.74 & 27.60 & 10.00 & 0.50 & 0.47 & 7.39 & 26.25 & 6.80 & 34.42 & 1.54 & 2.88 & 0.00 & 7.24 & 7.88 & 4.68 \\

H2GFormer-T~\cite{wang2024h2gformer} & 44.69 & 14.29 & 57.00 & 29.37 & \textbf{21.74} & 0.34 & 20.51 & 28.21 & 6.80 & 0.95 & 0.91 & 9.32 & 27.44 & 7.80 & 36.26 & 1.15 & 0.10 & 0.00 & 7.98 & 9.88 & 5.81 \\

HASSC-T~\cite{wang2024not} & 44.58 & 14.74 & 57.23 & 29.08 & 19.89 & \textbf{1.26} & 20.19 & 27.33 & 17.06 & 1.07 & 1.14 & 8.83 & 27.01 & 7.71 & 33.95 & 2.25 & 4.09 & 0.00 & 7.95 & 9.20 & 4.81 \\

Symphonies~\cite{jiang2024symphonize} & 41.92 & 14.89 & 56.37 & 27.58 & 15.28 & 0.95 & 21.64 & 28.68 & 20.44 & \textbf{2.54} & 2.82 & 13.89 & 25.72 & 6.60 & 30.87 & 3.52 & 2.24 & 0.00 & 8.40 & 9.57 & 5.76 \\

\hline

\shortname(Ours) & \textbf{56.56} & \textbf{20.54} & \textbf{67.90} & \textbf{37.04} & 20.13 & 0.02 & \textbf{32.53} & \textbf{44.04} & 15.16 & 0.67 & \textbf{2.86} & \textbf{17.37} & \textbf{39.57} & \textbf{17.51} & \textbf{47.45} & 2.77 & 1.18 & 0.00 & \textbf{13.63} & \textbf{22.14} & \textbf{8.28} \\
\hline
\end{tabular}
}
\label{tab:semanti_kitti_results}
\end{table*}
\begin{figure}[t]
\centering
    \includegraphics[width=1.0\linewidth]{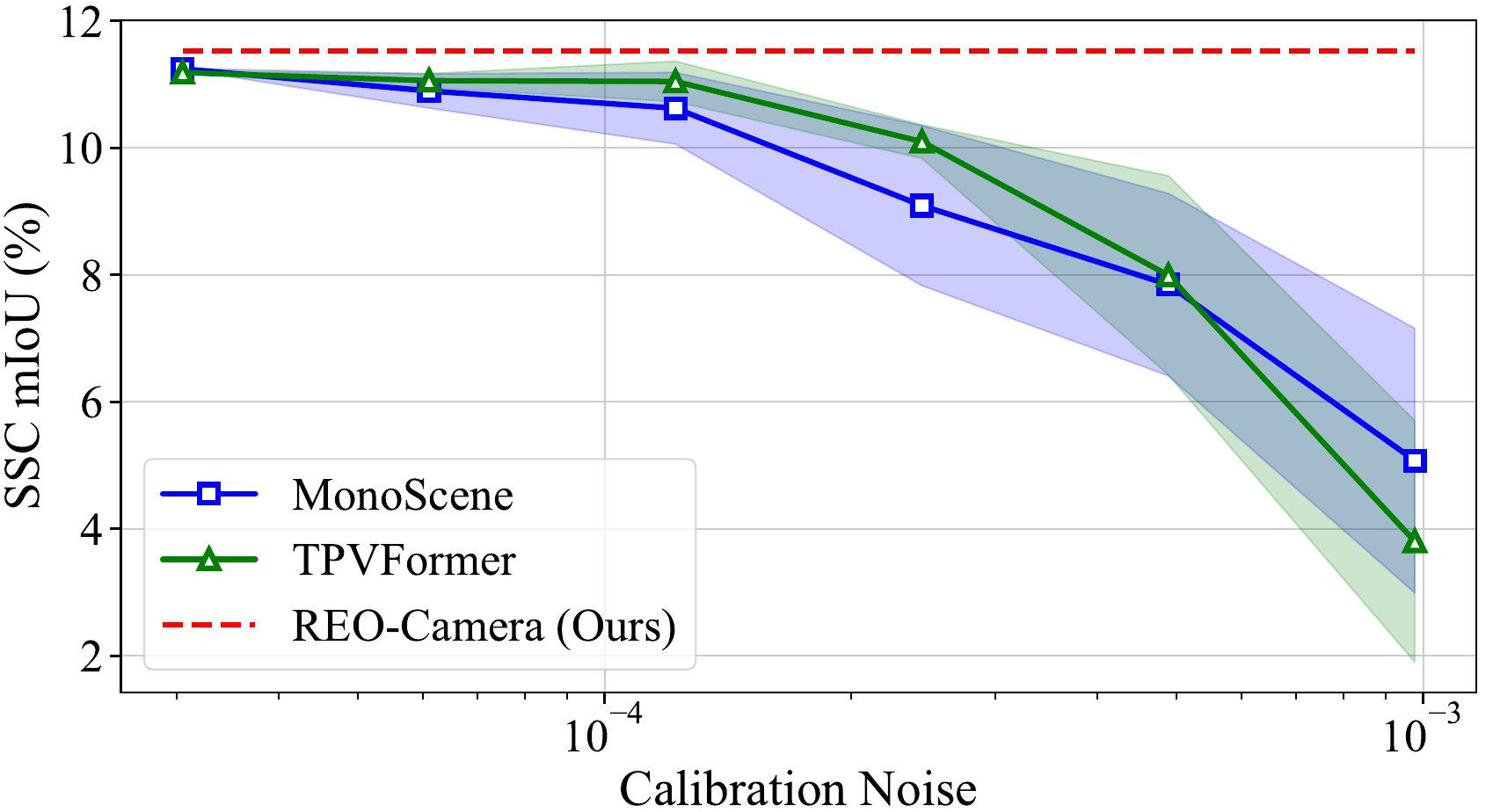}
    \caption{Comparisons of the model performance under different levels of calibration noise on SemanticKITTI Scene Completion.}
    \label{fig:calib_noise_analyze}
\end{figure}

\begin{figure*}
    \centering
    \includegraphics[width=1.0\linewidth]{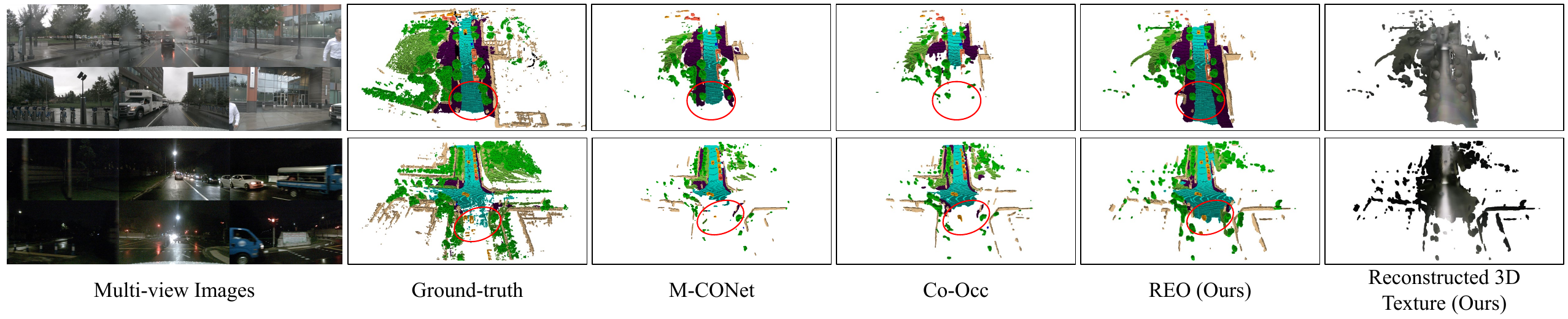}
    \caption{Qualitative results of \shortname on OpenOccupancy. We highlight the main differences with red circles. Better viewed by zooming in.}
    \label{fig:openocc_visualize}
\end{figure*}

\begin{table}
    \centering
    \caption{Effect of the number of voxels on SemanticKITTI.}
    \scalebox{0.93}{
    \begin{tabular}{c|ccccccc}
    \hline
    \#Point cloud voxels & 0 (Camera-only) & 2560 & 5120 & 10240 & 20480 % 40960 
    \\ \hline
    \#Params. (M)$\downarrow$ & 39.04 & 42.70 & 42.70 & 42.70 & 42.70
    \\
    \#FLOPs (G)$\downarrow$ & 178 & 206 & 216 & 234 & 271
    \\
    SC IoU (\%)$\uparrow$ & 35.34 & 51.95 & 53.42 & 55.34 & \textbf{56.56} \\
    SSC mIoU (\%)$\uparrow$ & 11.53 & 18.12 & 18.85 & 19.39 & \textbf{20.54} \\
    \hline
    \end{tabular}
    }
    \label{tab:effect_num_voxels}
\end{table}

\noindent\textbf{Results on SemanticKITTI Scene Completion.}
Following~\cite{huang2023tri, cao2022monoscene}, we further evaluate our method on SemanticKITTI Scene Completion and compare the performance to existing camera-centric methods, including TPVFormer~\cite{huang2023tri}, OccFormer~\cite{zhang2023occformer}, and Symphonies~\cite{jiang2024symphonize}. From Table~\ref{tab:semanti_kitti_results}, our \shortname outperforms existing camera-centric methods by a large margin in both SC IoU and SSC mIoU. Specifically, our \shortname surpasses the state-of-the-art method (\ie, Symphonies~\cite{jiang2024symphonize}) by 14.64\% and 5.65\% in SC IoU and SSC mIoU, respectively. 

As our \shortname fuses 20,480 voxels on SemanticKITTI, we investigate the effect of the number of fused point cloud voxels. From table~\ref{tab:effect_num_voxels}, fusing only a few numbers of point cloud voxels brings promising improvement to our \shortname. Specifically, fusing only 2560 voxels brings 16.61
\% and 6.59\% improvements in SC IoU and SSC mIoU. Notably, our \shortname with 2560 point cloud voxels outperforms the state-of-the-art camera-centric method (\ie, Symphonies~\cite{jiang2024symphonize}) by 10.03\% and 3.23\% in SC IoU and SSC mIoU.

\begin{table}
    \centering
    \caption{Comparisons of model efficiency on OpenOccupancy. $\dagger$ denotes the results with multi-sweep point clouds. The \textbf{bold} numbers denote the best results.}
    \scalebox{0.87}{
    \begin{tabular}{c|ccccc}
    \hline
    Method & \#Parmas.$\downarrow$ & \#FLOPs$\downarrow$  & Inference Time$\downarrow$ & mIoU$\uparrow$ & IoU$\uparrow$  \\ \hline
    C-CONet~\cite{wang2023openoccupancy} & 117.99M & 2371G & 394.77ms & 13.6\% & 21.6\% \\
    L-CONet$^\dagger$~\cite{wang2023openoccupancy} & 65.63M & 810G & 266.32ms & 15.9\% & 30.1\% \\
    M-CONet$^\dagger$~\cite{wang2023openoccupancy} & 125.34M & 3066G & 736.50ms & 20.5\% & 26.5\% \\
    % PointOcc~\cite{zuo2023pointocc} & 29.34M & 599G & 140.28ms & 20.5\% & 30.9\% \\
    % PointOcc$^\dagger$~\cite{zuo2023pointocc} & 29.34M & 625G & 151.33ms & 23.9\% & 34.1\% \\
    Co-Occ$^\dagger$~\cite{pan2024co} & 225.68M & -- & 1460.21ms & \textbf{21.9}\% & 30.6\% \\
    \hline
    \shortname (Ours) & \textbf{43.77}M & \textbf{397}G & \textbf{73.66}ms & 21.3\% & \textbf{31.7}\% \\
    % \shortname-R50$^\dagger$(Ours) & 43.77M & 508G &  & 21.2\% & 33.7\% \\
    % \shortname-R101(Ours) & 62.76M & 557G & 94.62ms &  21.4\% & 31.8\% \\
    \hline
    \end{tabular}
    }
    \label{tab:analysis_efficient}
\end{table}

\begin{table}[t]
    \centering
    \caption{Detailed inference time of each component in \shortname. We report the results on SemanticKITTI. }
    \scalebox{0.90}{
    \begin{tabular}{c|cccccc}
    \hline
    Components & Backbone & Image2BEV & PCD2BEV & Decoder & Total \\
    \hline
    % Cost Time  & 7.30ms & 5.02ms & 19.60ms & 8.71ms & 3.61ms & 44.26ms \\ 
    Cost Time $\downarrow$ & 7.30ms & 5.02ms & 19.60ms & 12.34ms & 44.26ms \\ 
    \hline
    \end{tabular}
    }
    \label{tab:detailed_time_smk}
\end{table}

\subsection{Efficiency analysis}
We analyze the efficiency of our method using a single GeForce RTX 3090 and compare it with existing methods on OpenOccupancy. From Table~\ref{tab:analysis_efficient}, our \shortname shows promising efficiency compared with existing fusion-based methods. For instance, our \shortname achieves 19.8$\times$ acceleration compared to Co-Occ~\cite{pan2024co}, with 1.1\% improvements in geometry IoU. Moreover, our \shortname also outperforms M-CONet~\cite{wang2023openoccupancy} by 5.2\% and 0.8\% in geometry IoU and semantic mIoU, respectively, with 10.0$\times$ speedup. These results demonstrate that our method is efficient enough for real-time applications like autonomous driving.

To investigate the efficiency of the proposed calibration-free spatial transformation module, we further report the cost time of spatial transformation \wrt single-view image and compare it with the baselines including VoxFormer~\cite{li2023voxformer}, TPVFormer~\cite{huang2023tri}, and MonoScene~\cite{cao2022monoscene} on SemanticKITTI. As shown in Tables~\ref{tab:dca_infer_speed} and~\ref{tab:detailed_time_smk}, our CST only takes 11.34\% of the total cost time and shows promising efficiency compared to existing methods.

\begin{table}[t]
    \centering
    \caption{Robustness of \shortname against calibration noise on SemanticKITTI.}
    % \vskip -0.1in
    \scalebox{1.0}{
    \begin{tabular}{c|cc}
    \hline
    Settings & SC IoU (\%)$\uparrow$ & SSC mIoU (\%)$\uparrow$\\
    \hline
    \shortname w/o calibration noise & 56.56 & 20.54 \\
    \shortname with calibration noise & 56.06 & 20.43 \\
    \hline
    \end{tabular}
    }
    \label{tab:calib_noise}
    % \vskip -0.1in
\end{table}

\begin{table*}
\centering
\caption{LiDAR segmentation results on nuScenes validation set. Cons. Veh represents construction vehicle and Dri. Sur denotes driveable surface, respectively. The \textbf{bold} numbers indicate the best results.} 
% \vskip 0.15in
 % scalebox{0.6}
% \renewcommand\arraystretch{1.1}
  \scalebox{0.95}{
\begin{tabular}{l|c|cccccccccccccccc}
\hline
Method 
% & Backbone
% & \rotatebox{90}{mIoU (\%)} 
& mIoU (\%)
& \rotatebox{90}{barrier} 
& \rotatebox{90}{bicycle} 
& \rotatebox{90}{bus} 
& \rotatebox{90}{car} 
& \rotatebox{90}{Cons. Veh} 
& \rotatebox{90}{motorcycle} 
& \rotatebox{90}{pedestrian} 
& \rotatebox{90}{traffic cone} 
& \rotatebox{90}{trailer} 
& \rotatebox{90}{truck} 
& \rotatebox{90}{Dri. Sur} 
& \rotatebox{90}{other flat} 
& \rotatebox{90}{sidewalk} 
& \rotatebox{90}{terrain} 
& \rotatebox{90}{manmade} 
& \rotatebox{90}{vegetation} 
\\ 

\hline\hline
% \hline
BEVFormer~\cite{li2022bevformer} & 56.2 & 54.0 & 22.8 & 76.7 & 74.0 & 45.8 & 53.1 & 44.5 & 24.7 & 54.7 & 65.5 & 88.5 & 58.1 & 50.5 & 52.8 & 71.0 & 63.0 \\
OccNet~\cite{tong2023scene} & 60.5 & 67.0 & 32.6 & 77.4 & 73.9 & 37.6 & 50.9 & 51.5 & 33.7 & 52.2 & 67.1 & 88.7 & 58.0 & 58.0 & 63.1 & 78.9 & 77.0 \\
TPVFormer~\cite{huang2023tri}  & 68.9 & 70.0 & 40.9 & 93.7 & 85.6 & 49.8 & \textbf{68.4} & 59.7 & 38.2 & 65.3 & 83.0 & 93.3 & 64.4 & 64.3 & 64.5 & 81.6 & 79.3 \\
OccFormer~\cite{zhang2023occformer} & 70.4 & 70.3 & 43.8 & 93.2 & 85.2 & 52.0 & 59.1 & 67.6 & 45.4 & 64.4 & \textbf{84.5} & 93.8 & 68.2 & 67.8 & 68.3 & 82.1 & 80.4\\
PanoOcc~\cite{wang2024panoocc} & 71.6 & 74.3 & 43.7 & 95.4 & 87.0 & \textbf{56.1} & 64.6 & 66.2 & 41.4 & \textbf{71.5} & 85.9 & 95.1 & 70.1 & 67.0 & 68.1 & 80.9 & 77.4 \\
% PointOcc~\cite{zuo2023pointocc} & 77.9 & 78.3 & 44.5 & 92.6 & 92.2 & 56.4 & 83.6 & 80.5 & 65.2 & 69.0 & 82.4 & 97.0 & 75.0 & 76.3 & 75.1 & 90.1 & 87.8 \\
\hline

\shortname (Ours) & \textbf{73.2} & \textbf{75.0} & \textbf{44.5} & \textbf{95.5} & \textbf{92.6} & 53.6 & 55.4 & \textbf{69.0} & \textbf{48.9} & 70.6 & 84.9 & \textbf{95.4} & \textbf{73.0} & \textbf{71.1} & \textbf{71.9} & \textbf{85.7} & \textbf{83.7} \\
\hline
\end{tabular}
}
\label{tab:nus_lidar_seg}
\end{table*}

\begin{figure*}
    \centering
    \includegraphics[width=1.0\linewidth]{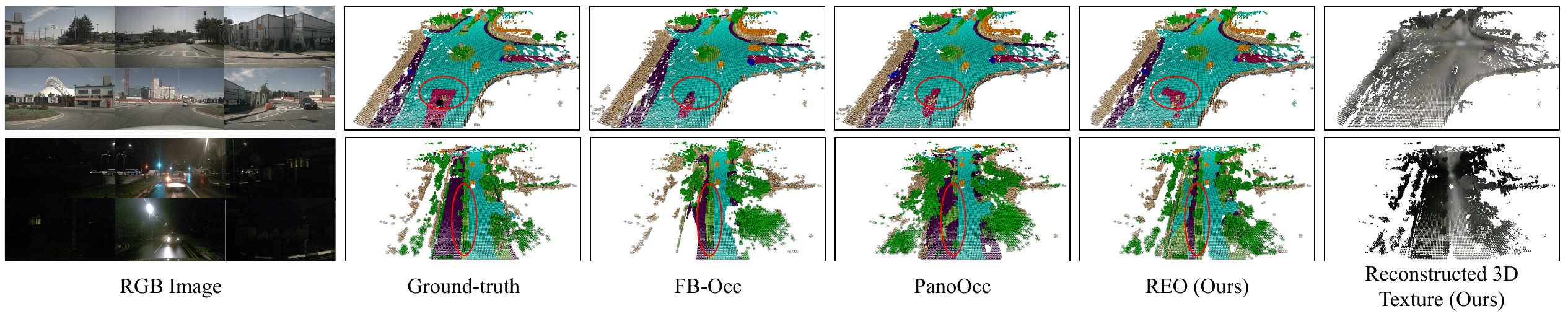}
    \caption{Qualitative results of \shortname on Occ3D-nuScenes. We highlight the main differences with red circles. Better viewed by zooming in.}
    \label{fig:occ3d_nus_visualize}
\end{figure*}

\begin{figure*}
    \centering
    \includegraphics[width=1.0\linewidth]{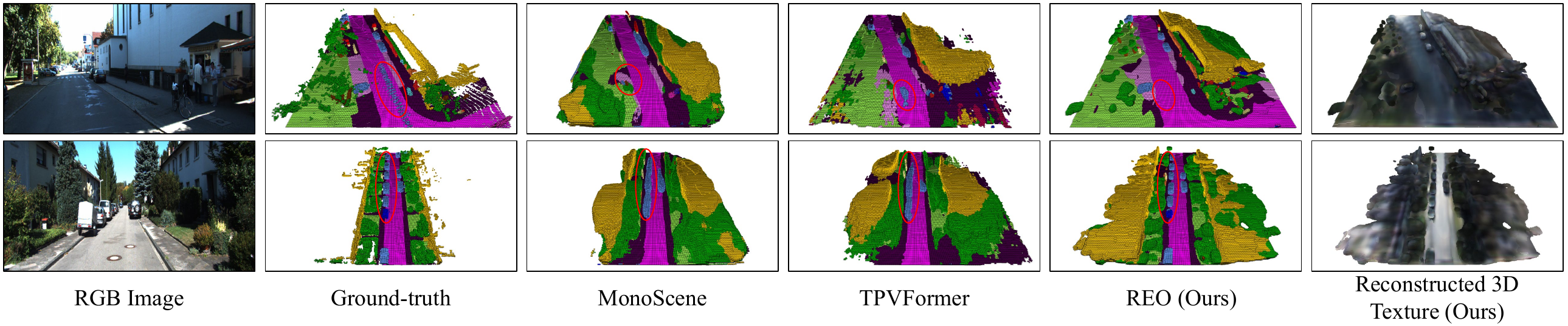}
    \caption{Qualitative results of \shortname on SemanticKITTI Scene Completion. We highlight the differences with red circles. Better viewed by zooming in.}
    \label{fig:smk_visualize}
\end{figure*}
\subsection{Robustness against calibration noise}
\label{sec:discuss_calibration}
By introducing the calibration-free spatial transformation, our \shortname can implicitly learn the 2D-to-3D spatial projection from multi-view 2D images. In this way, our method avoids the use of sensor calibration during inference. In contrast, existing methods that rely on sensor calibration may be affected by the calibration noise in practice. To verify this, we apply different levels of Gaussian noise to sensor calibration. Specifically, given the extrinsic of camera $\bR\in \mmR^{4 \times 4}$, we adopt the noise by $\bR_{noise} =\bR + e$, where $e\sim N(\mu, \sigma^2)$, $\mu=0$ and $\sigma\in \{2^{-15},2^{-14}, 2^{-13}, 2^{-12}, 2^{-11}, 2^{-10} \}$. As shown in Figure~\ref{fig:calib_noise_analyze}, as existing methods rely on sensor calibration, their model performance is degraded with the increase of calibration noise. In contrast, we do not use the sensor calibration and the spatial projection is learned implicitly with the proposed CST. Therefore, our method is robust to the changes in sensor calibration.

Since our method also uses sensor calibration to build the ground truth of occupancy prediction and 2D/3D auxiliary training tasks, we further investigate the impact of calibration noise during training. Specifically, we apply calibration noise with $\sigma=2^{-10}$ ($\approx 10^{-3}$) to the camera extrinsic when constructing ground truth during training. Note that existing methods suffer from significant performance degradation under the same noise level. From Table~\ref{tab:calib_noise}, \shortname shows promising robustness against calibration noise, as it constructs the 3D space from a larger attention region rather than a single reference point like existing spatial transformation methods.

\subsection{Qualitative evaluation}
To better understand the benefits of our \shortname, we visualize the occupancy prediction results on the three benchmarks. Specifically, on OpenOccupancy, we compare our \shortname with the state-of-the-art fusion-based methods,~\ie, M-CONet~\cite{wang2023openoccupancy}, and Co-Occ~\cite{pan2024co}. As demonstrated in Figure~\ref{fig:openocc_visualize}, our \shortname performs better on geometry occupancy prediction compared to the baselines. On Occ3D-nuScenes, we compare our \shortname with FB-Occ~\cite{li2023fb} and PanoOcc~\cite{wang2024panoocc}. From Figure~\ref{fig:occ3d_nus_visualize}, all the methods show similar performance at day-time. While at night-time, our \shortname shows superior performance on semantic occupancy prediction by fusing the spatial information from LiDAR. On SemanticKITTI Scene Completion, we compare our \shortname with two camera-only baselines, \ie, MonoScene~\cite{cao2022monoscene}, TPVFormer~\cite{huang2023tri}. As demonstrated in Figure~\ref{fig:smk_visualize}, our \shortname shows better scene completion results over the baselines by fusing both information from the camera and LiDAR. Besides, we notice that the ground truth may contain some noise as it is constructed by stacking multi-sweep point clouds. In this case, our \shortname has better robustness against the noise labels compared to TPVFormer.
% To better understand the benefits of our \shortname, we visualize scene completion results on SemanticKITTI in Figure~\ref{fig:smk_visualize}. Compared to the existing methods (\ie, MonoScene~\cite{cao2022monoscene}, TPVFormer~\cite{huang2023tri}), our method achieves better performance on scene reconstruction by fusing multi-modal information from different sensors. In addition, we visualize the semantic occupancy prediction of our \shortname under different modalities. From Figure~\ref{fig:nus_visualize}, our \shortname-Camera reconstructs the surroundings from multi-view images without using sensor calibration information. However, it failed to detect the small object like a pole. In comparison, our \shortname performs better by incorporating spatial information from LiDAR.

\subsection{Comparisons on LiDAR semantic segmentation}
By replacing the queried voxels with point clouds, our \shortname can be easily applied to the LiDAR segmentation task. Therefore,  we further evaluate the performance of our \shortname on nuScenes LiDAR segmentation. Following the settings on OpenOccupancy, we use ResNet-50 as the backbone of the 2D image encoder and train the model for 25 epochs with a batch size of 16. As shown in Table~\ref{tab:nus_lidar_seg}, our \shortname achieves the best performance over the existing occupancy prediction methods, such as OccFormer~\cite{zhang2023occformer} and PanoOcc~\cite{wang2024panoocc}. Specifically, our \shortname outperforms PanoOcc~\cite{wang2024panoocc} by 1.6\% in mIoU.  

\begin{table}
    \centering
    \caption{Ablation study of the proposed network components on Occ3D-nuScenes validation set. \textbf{IFA} denotes image feature aggregation, \textbf{BCB} denotes BEV contextual blocks, \textbf{Auxiliary Tasks} denotes the proposed 2D/3D auxiliary training tasks. The \textbf{bold} numbers denote the best results.}
    \scalebox{0.94}{
    \begin{tabular}{ccccc|cc}
    \hline
    & Baseline & IFA & BCB & Auxiliary Tasks & IoU (\%)$\uparrow$ & mIoU (\%)$\uparrow$ \\
    \hline
    1 & \checkmark & & & & 80.51 & 42.40\\
    \hdashline
    2 & \checkmark & \checkmark & & & 80.81 & 43.04 \\
    3 & \checkmark & \checkmark & \checkmark &  & \textbf{81.97} & 42.68 \\
    4 & \checkmark & \checkmark & & \checkmark & 81.22 & 46.00 \\
    5 & \checkmark & \checkmark & \checkmark & \checkmark  & 81.67 & \textbf{46.22}\\
    \hline
    \end{tabular}
    }
    \label{tab:abl_components}
\end{table}

\begin{table}[t]
    \centering
    \caption{Effect of using different stages of image features on Occ3D-nuScenes validation set. The \textbf{bold} numbers indicate the best results.}
    \scalebox{0.96}{
    \begin{tabular}{ccccc|cc}
    \hline
    & Stage-1 & Stage-2 & Stage-3 & Stage-4 & IoU (\%)$\uparrow$ & mIoU (\%)$\uparrow$ \\
    \hline
    1 & \checkmark & & & & 80.87 & 39.11 \\
    2 &  & \checkmark & & & \textbf{81.91} & 41.69 \\
    % 3 & & &  \checkmark & \checkmark & 46.09\\
    3 & & & \checkmark & & 81.74 & 45.37 \\
    4 & & & & \checkmark & 81.67 & \textbf{46.22}\\
    5 & \checkmark & \checkmark & \checkmark & \checkmark & 81.90 & 46.21\\
    % 5 &  & \checkmark & \checkmark & \checkmark & 46.17 \\
    % 6 & \checkmark &  & \checkmark & \checkmark & 46.37 \\
    % 7 & \checkmark & \checkmark &  & \checkmark & 45.93 \\
    % 8 & \checkmark & \checkmark & \checkmark & & 44.85 \\
    % 9 & \checkmark & \checkmark & \checkmark & \checkmark & \textbf{46.22} \\
    \hline
    \end{tabular}
    }
    \label{tab:abl_multi_level_agg}
\end{table}

\begin{figure}[htp]
    \centering
    \includegraphics[width=0.95\linewidth]{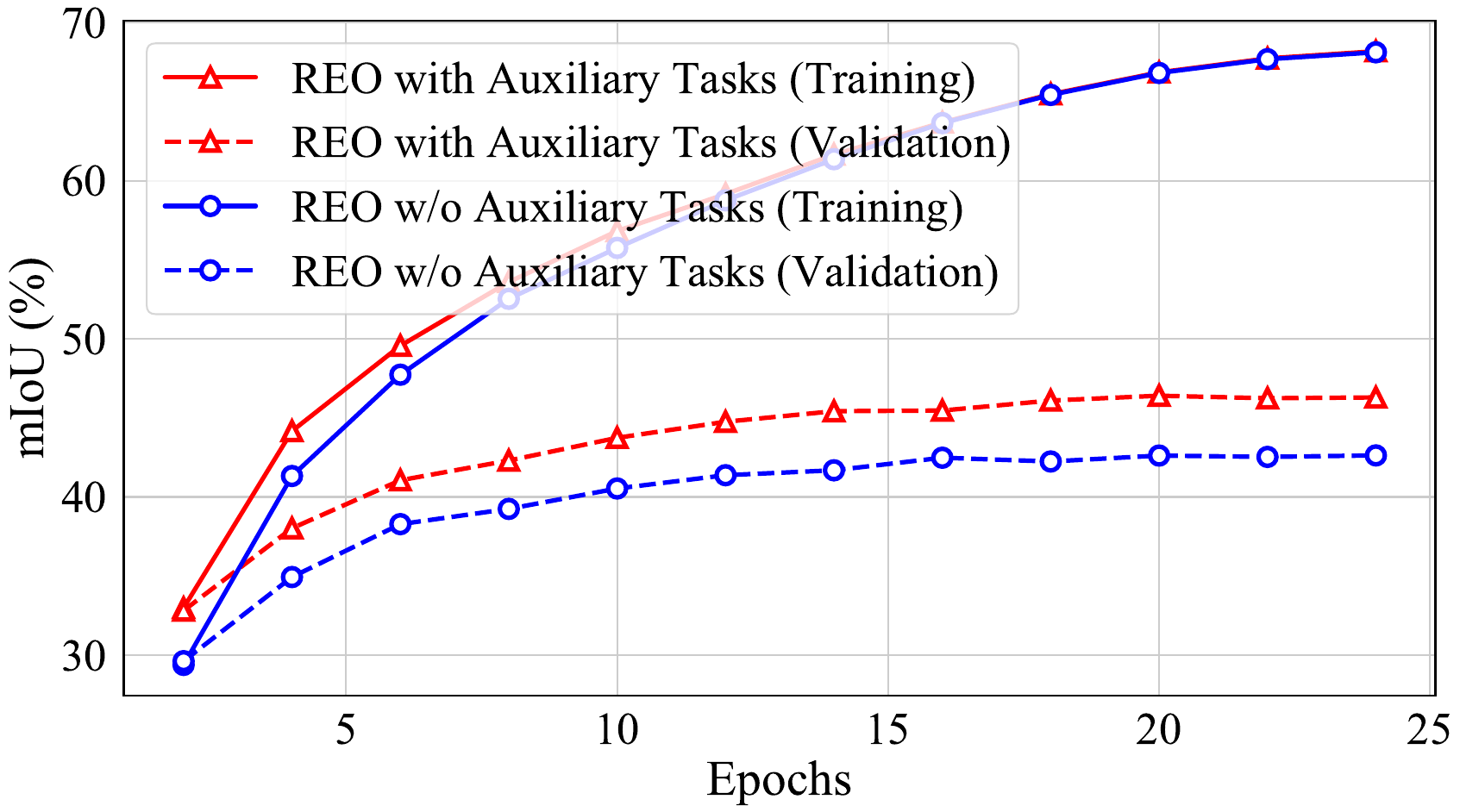}
    \caption{Training and validation mIoU of \shortname w/wo auxiliary training tasks on Occ3D-nuScenes.}
    \label{fig:trainval_curve_occ3d}
\end{figure}

\begin{figure*}
    \centering
    \includegraphics[width=0.98\linewidth]{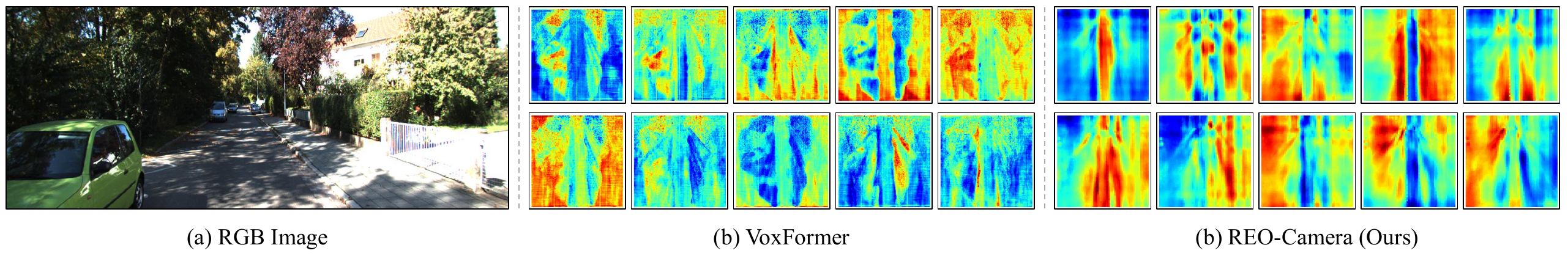}
    \caption{Visualization results of the projected BEV features. For clarity, we only show the first 10 BEV features of both VoxFormer and \shortname on SemanticKITTI. Better viewed by zooming in.}
    \label{fig:smk_bev_feat}
\end{figure*}
\section{Ablation study}
\subsection{Effect of the proposed network components}
We study the effect of the proposed network components of \shortname on Occ3D-nuScenes validation set, including image feature aggregation (IFA), BEV contextual block (BCB), and 2D/3D auxiliary training tasks. Note that we replace the proposed SFA and BCB with a single convolutional layer when investigating the effect of the proposed modules. The hyper-parameters of objectives are empirically set to 1.0. As shown in Table~\ref{tab:abl_components}, the proposed image feature aggregation brings 0.30\% and 0.64\% improvements in mIoU and IoU, respectively. By comparing the second and third rows, using BEV contextual block improves the IoU by 1.16\%. Moreover, comparing the third and fifth rows, the proposed auxiliary training tasks bring 3.57\% mIoU improvement to the occupancy network. In addition, as demonstrated in Figure~\ref{fig:trainval_curve_occ3d}, the proposed auxiliary training tasks effectively help the model learn discriminative features from multi-view images, resulting in better generalization ability on the validation set.

In our spatial-aware feature aggregation module, we fuse the fourth-stage features from the 2D image backbone. As shown in Table~\ref{tab:abl_multi_level_agg}, we further conduct experiments to investigate the effect of using different stages of features. From the results, using the fourth-stage features from the image backbone achieve the best performance in mIoU. Besides, we also resize the image features to the sample resolution and merge them with a concatenation layer to obtain the performance of using all features. By comparing the fourth and fifth rows, using all stage features only brings marginal improvements in IoU. To trade off the model performance and efficiency, we only use the fourth-stage features from the image backbone in our main experiments.

\subsection{Effect of hyperparameters in objectives}
In this section, we investigate the effect of hyperparameters used in the objectives. First, we set $\lambda_{dice},\lambda_{depth},\lambda_{rgb}$ to 1.0 and train our \shortname with $\lambda_{focal}\in \{0, 0.5, 1.0, 1.5, 2.0\}$ on Occ3D-nuScenes. From Table~\ref{tab:eff_focal_loss}, using focal loss significantly improves the semantic classification power of networks. Besides, the model performance is not sensitive to $\lambda_{focal}$ when increasing $\lambda_{focal}$ from 0.5 to 2.0. Second, we set $\lambda_{focal}, \lambda_{depth}, \lambda_{rgb}$ to 1.0 and train the model with $\lambda_{dice}\in \{0, 0.5, 1.0, 1.5, 2.0\}$. As shown in Table~\ref{tab:eff_dice_loss}, using dice loss can effectively improve the model performance in both geometry IoU and semantic mIoU. However, with the increasing of $\lambda_{dice}$, performance in semantic mIoU is degraded while the performance in geometry occupancy prediction is slightly improved. To trade off geometry and semantic occupancy prediction performance, we use $\lambda_{dice}=1$ in our main experiments. Third, we set $\lambda_{focal},\lambda_{dice},\lambda_{rgb}$ to 1 and train the model with $\lambda_{depth} \in \{0, 1.0, 2.0, 4.0, 8.0\}$. From Table~\ref{tab:eff_depth_loss}, the model performance in semantic occupancy prediction can be improved by introducing a depth regression task. We then use $\lambda_{depth}=2.0$ and train model with $\lambda_{rgb}\in \{0, 1.0, 2.0, 4.0, 8.0\}$ to study the effect of the texture reconstruction task. As shown in Table~\ref{tab:eff_rgb_loss}, using texture reconstruction can further improve the model performance in mIoU.

\begin{table}[t]
    \centering
    \caption{Effect of $\lambda_{focal}$. The \textbf{bold} numbers denote the best results.}
    \begin{tabular}{c|ccccc}
    \hline
    $\lambda_{focal}$ & 0 & 0.5 & 1.0 & 1.5 & 2.0 \\
    \hline
    IoU (\%)$\uparrow$ & 81.32 & 81.23 & \textbf{81.67} & 81.47 & 81.43 \\
    mIoU (\%)$\uparrow$ & 44.25 & 45.63 & \textbf{46.22} & 46.12 & 45.96 \\
    \hline
    \end{tabular}
    \label{tab:eff_focal_loss}
\end{table}

\begin{table}[t]
    \centering
    \caption{Effect of $\lambda_{dice}$. The \textbf{bold} numbers denote the best results.}
    \begin{tabular}{c|ccccc}
    \hline
    $\lambda_{dice}$ & 0 & 0.5 & 1.0 & 1.5 & 2.0\\
    \hline
    IoU (\%)$\uparrow$ & 80.88 & 81.21 & 81.67 & 81.71 & \textbf{81.72} \\
    mIoU (\%)$\uparrow$ & 42.72 & \textbf{46.34} & 46.22 & 46.05 & 45.43 \\
    \hline
    \end{tabular}
    \label{tab:eff_dice_loss}
\end{table}

\begin{table}[t]
    \centering
    \caption{Effect of $\lambda_{depth}$. The \textbf{bold} numbers denote the best results.}
    \begin{tabular}{c|ccccc}
    \hline
    % $\lambda_{depth}$ & 0 & 0.5 & 1.0 & 1.5 & 2.0 & 3.0 & 4.0 \\
    % \hline
    % IoU (\%) & 81.71 & 81.52 & 81.67 & 81.65 & 81.91 & 81.61 \\
    % mIoU (\%) & 46.11 & 45.94 & 46.22 & 46.25 & \textbf{46.51} & 46.17\\
    $\lambda_{depth}$ & 0 & 1.0 & 2.0 & 4.0 & 8.0 \\
    \hline
    IoU (\%)$\uparrow$ & 81.88 & 81.67 & 81.91 & 81.70 & \textbf{82.11} \\
    mIoU (\%)$\uparrow$ & 45.94 & 46.22 & \textbf{46.51} & 46.30 & 46.11 \\
    \hline
    \end{tabular}
    \label{tab:eff_depth_loss}
\end{table}

\begin{table}[t]
    \centering
    \caption{Effect of $\lambda_{rgb}$. The \textbf{bold} number denotes the best result.}
    \begin{tabular}{c|cccccc}
    \hline
    $\lambda_{rgb}$ & 0 & 1.0 & 2.0 & 4.0 & 8.0 \\
    \hline
    IoU (\%)$\uparrow$ & 81.70 & \textbf{81.91} & 81.72 & 81.15 & 81.44 \\
    mIoU (\%)$\uparrow$ & 45.85 & 46.51 & 46.57 & 46.60 & \textbf{46.86} \\
    \hline
    \end{tabular}
    \label{tab:eff_rgb_loss}
\end{table}

\begin{table}[t]
    \centering
    \caption{Effect of using different image backbones on Occ3D-nuScenes validation set. The \textbf{bold} numbers indicate the best results.}
    \begin{tabular}{c|ccccc}
    \hline
    Backbones & R-18 & R-34 & R-50 & R-101 & R-152 \\
    \hline
    \#Params. (M)$\downarrow$ & 28.28 & 38.39 & 41.21 & 60.20 & 75.84 \\
    \#FLOPs (G)$\downarrow$ & 186 & 265 & 285 & 444 & 604 \\
    IoU (\%)$\uparrow$ & 81.55 & 81.78 & \textbf{81.91} & 81.61 & 81.55\\
    mIoU (\%)$\uparrow$ & 44.72 & 45.87 & 46.51 & 46.59 & \textbf{46.72} \\
    \hline
    \end{tabular}
    \label{tab:abl_backbone}
\end{table}

\subsection{Effect of using different image backbones}
We investigate the effect of using different image backbones, \ie, ResNet-18, ResNet-34, ResNet-50, ResNet-101, and ResNet-152 on Occ3D-nuScenes validation set. As shown in Table~\ref{tab:abl_backbone}, using deeper networks can bring performance gains to our method. For example, using ResNet-50 as backbones outperforms the model with ResNet-34 by 0.36\% and 1.79\% in geometry IoU and semantic mIoU, respectively. Note that deeper networks also require more computation resources. Therefore, we use ResNet-50 as the backbone to extract the image features from multi-cameras.

\subsection{Effect of different resolutions of image}
We further study the effect of using different resolutions of RGB images as inputs. Specifically, we train the models with resolutions of $224\times 416$, $448\times 800$, and $896\times 1600$, and report the results on Occ3D-nuScenes validation set. As shown in Table~\ref{tab:abl_img_resolution}, using a higher resolution of input images can further improve the model performance. However, it also leads to higher computational complexity of the model. For instance, the model with a resolution of $896\times 1600$ requires 3.0$\times$ of computational costs more than the one with a resolution of $448\times 800$. 

\begin{table}[t]
    \centering
    \caption{Effect of different resolutions of the image. The \textbf{bold} numbers indicate the best results}
    \begin{tabular}{c|cccc}
    \hline
    Resolution & \#Params.(M)$\downarrow$ & \#FLOPs(G)$\downarrow$ & IoU (\%)$\uparrow$ & mIoU (\%)$\uparrow$ \\
    \hline
    224$\times$416  & 40.51 & 144 & 81.72 & 43.60 \\
    448$\times$800  & 41.21 & 285 & 81.91 & 46.51 \\
    896$\times$1600 & 44.03 & 863 & \textbf{82.25} & \textbf{47.60} \\
    \hline
    \end{tabular}
    \label{tab:abl_img_resolution}
\end{table}

\subsection{Visualization of the projected BEV features}
\label{sec:vis_bev_feat}
To investigate the 2D-to-3D spatial projection learned by our calibration-free spatial transformation, we visualize the BEV feature maps of our \shortname on SemanticKITTI and compare the results to VoxFormer~\cite{li2023voxformer} for better illustration. For fair comparisons, we visualize the features from REO-Camera that use RGB images as inputs only. As shown in Figure~\ref{fig:smk_bev_feat}, our \shortname shows similar feature patterns as VoxFormer. These results prove that our CST can learn spatial correspondence even without using calibration information as inputs. Besides, we also observe that the features learned by our CST have better spatial continuity compared to those of VoxFormer, as our CST uses vanilla cross attention to model the spatial correspondence globally. 
% \subsection{Visualization of BEV feature maps}
% \label{sec:visualize_attn}
% To investigate the implicit spatial projection learned by our \shortname, we visualize the attention map of \shortname-Camera~\wrt different queried points on SemanticKITTI. Note that we use an empty image as input to avoid the influence of the image and focus on the 2D-to-3D projection learned by the model. As shown in Figure~\ref{fig:vis_attn_xy}, the attention of the model on the image is moving from left to right with the decrease of $y$-value of the queried points. Besides, with the increase of $x$-value of the queried points, the attention of model on the image is also moving from near to far. Moreover, we also investigate the changing of attention map~\wrt the queried points with different $z$-values. As shown in Figure~\ref{fig:visualize_attn_z}, the attention on the image is moving from bottom to top with the increase of $z$-value. By comparing Figure~\ref{fig:visualize_attn_z} (a) and Figure~\ref{fig:visualize_attn_z} (b), we observe that the attention map changes less with the increase in distance (\ie, with $x$-value changes from 5 to 10), which conforms to the fact that objects appear larger in the near distance and smaller in the far distance. In summary, these visualization results prove that our \shortname can effectively learn the 2D-to-3D spatial projection with our proposed 3D texture cues.

\section{Conclusion}
In this work, we proposed the calibration-free spatial transformation for robust and efficient semantic occupancy prediction. Different from the existing methods that explicitly use sensor calibration information to project the 2D RGB images to 3D space, we exploited a calibration-free projection method based on the vanilla cross-attention scheme and introduced a set of 2D/3D auxiliary training tasks to enhance the discrimination power of image backbones on both spatial and semantic features. Besides, our method is flexible to fuse different modalities to achieve better performance. The experimental results on three benchmarks showed the superiority of our method on both the robustness against calibration-noise and model efficiency for real-time application. In the future, we will fuse more modalities, including Radar and infrared cameras to further improve the model performance. In addition, we will try to apply our method to more perception tasks, such as road structure perception.

% use section* for acknowledgment
% \vspace{-0.1in}
% \ifCLASSOPTIONcompsoc
%   % The Computer Society usually uses the plural form
%   \section*{Acknowledgments}
% \else
%   % regular IEEE prefers the singular form
%   \section*{Acknowledgment}
% \fi

% \newpage
\ifCLASSOPTIONcaptionsoff
  \newpage
\fi
\bibliographystyle{abbrv}
{
	\bibliography{longstrings,reference}
}
% \vspace{-0.15in}
% \clearpage
\begin{IEEEbiography}[{\includegraphics[width=1in,height=1.25in,clip,keepaspectratio]{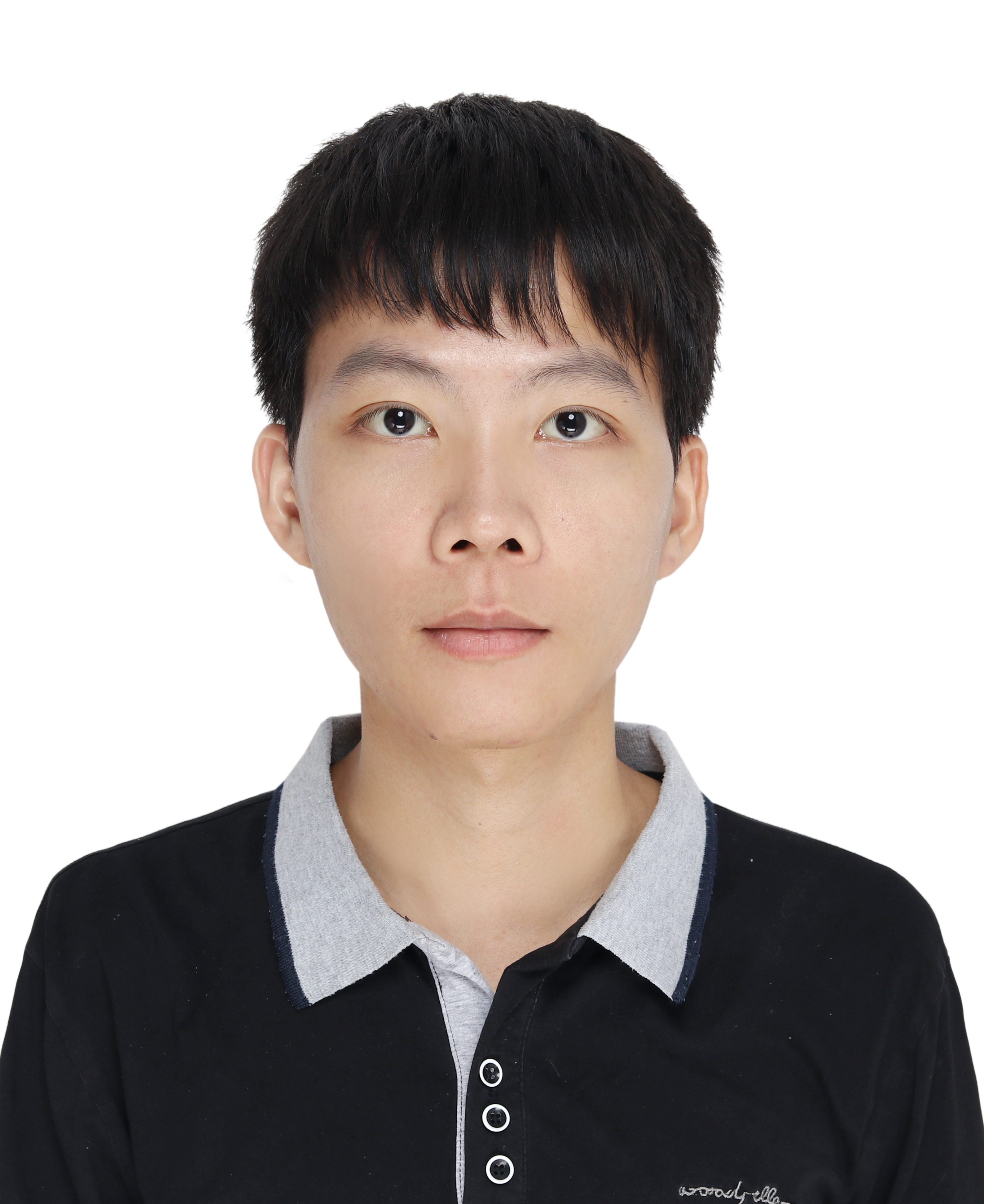}}]{Zhuangwei Zhuang}
is a Ph.D. student in the School of Software Engineering at South China University of Technology, and is currently working as an intern at RoboSense Inc., Shenzhen, China. He received his Bachelor Degree in Automation and Engineering in 2016 and Master Degree in Software Engineering in 2018, both from South China University of Technology in Guangzhou, China. His research interests include model compression and 3D scene understanding for autonomous driving.
\end{IEEEbiography}
\vskip -0.3in

\begin{IEEEbiography}[{\includegraphics[width=1in,height=1.25in,clip,keepaspectratio]{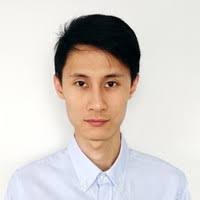}}]{Ziyin Wang}
is currently working as an AI director at RoboSense Inc., Shenzhen, China. He received the Bachelor Degree in 2013 at the School of Electrical Engineering from the East China University of Science and Technology, Shanghai, China. In 2019, he received the Ph.D. degree at the Department of Computer Science, Purdue University, West Lafayette, IN, USA. His research interests include autonomous driving, machine learning, and image understanding.
\end{IEEEbiography}
\vskip -0.3in

\begin{IEEEbiography}[{\includegraphics[width=1in,height=1.25in,clip,keepaspectratio]{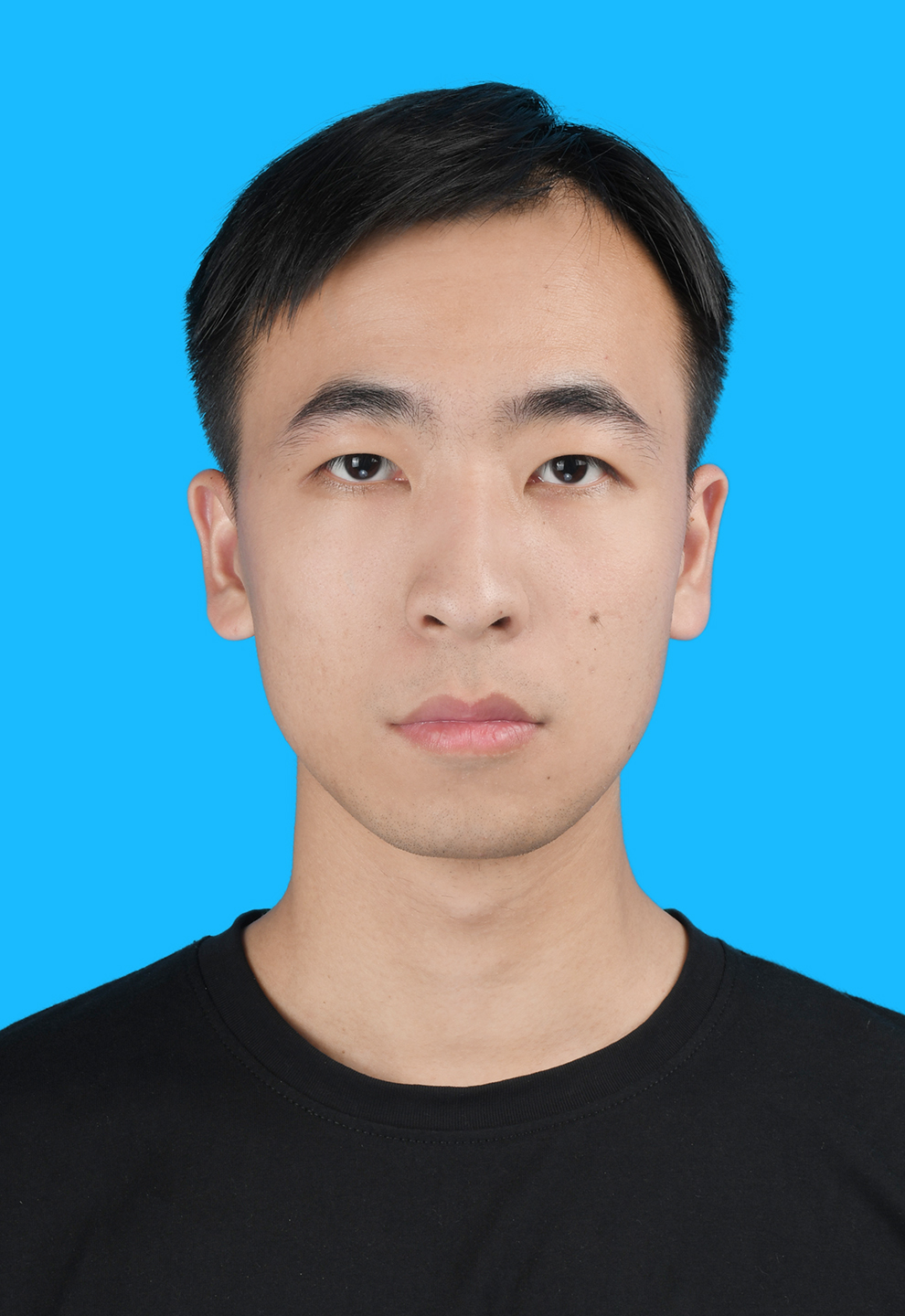}}]{Sitao Chen}
is a Master student in the School of Software Engineering at South China University of Technology.  He received his Bachelor Degree in the School of Mechanical \& Automotive in 2023 from South China University of Technology in Guangzhou, China. His research interests include 3D scene understanding for autonomous driving.
\end{IEEEbiography}
\vskip -0.3in

\begin{IEEEbiography}[{\includegraphics[width=1in,height=1.25in,clip,keepaspectratio]{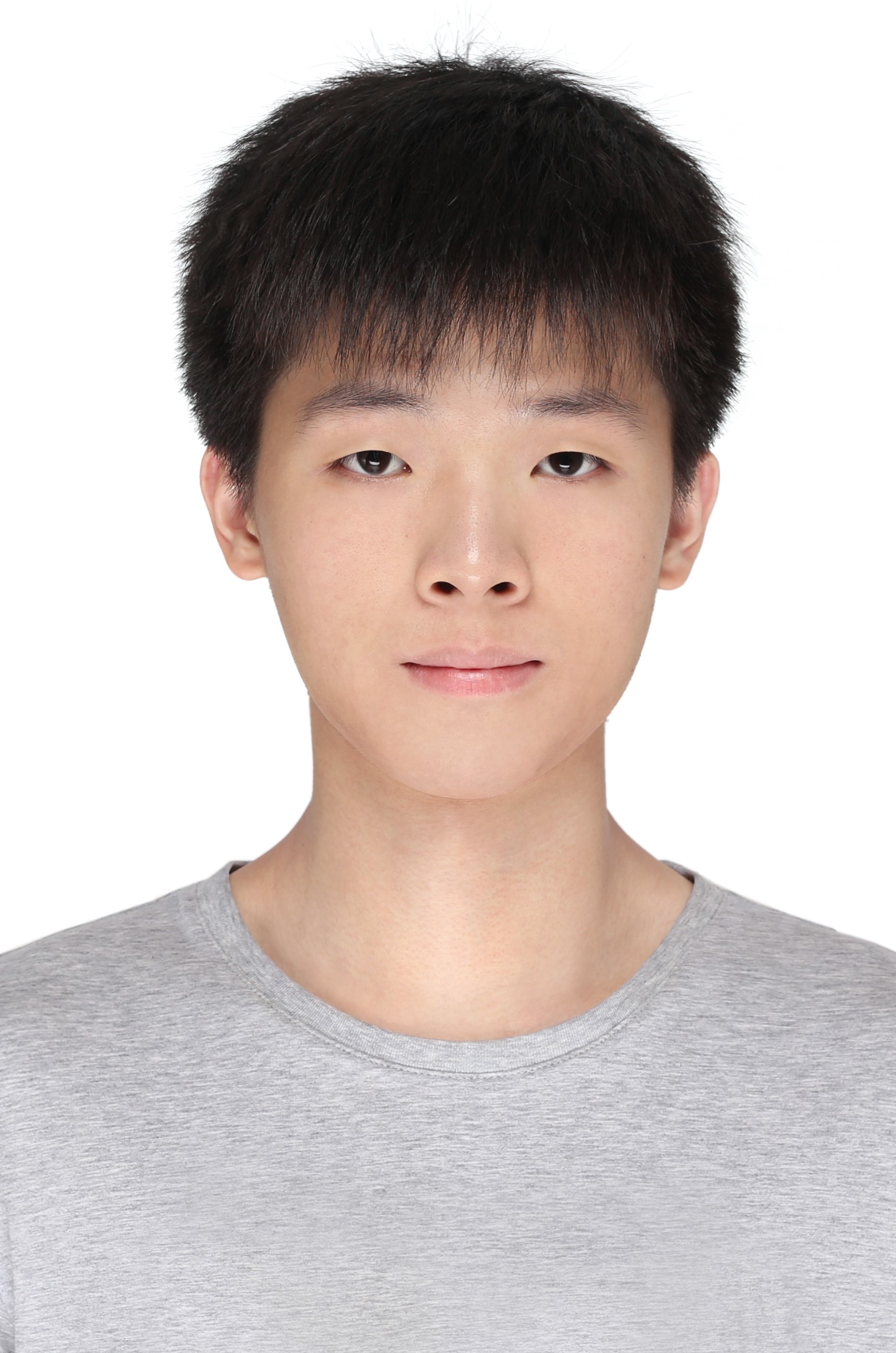}}]{Lizhao Liu}
is a Master student in the School of Software Engineering at South China University of Technology. He received his Bachelor Degree in the School of Software Engineering in 2021 from South China University of Technology in Guangzhou, China. His main research interests cover visual
recognition and multimodal learning.
\end{IEEEbiography}
\vskip -0.3in

\begin{IEEEbiography}[{\includegraphics[width=1in,height=1.25in,clip,keepaspectratio]{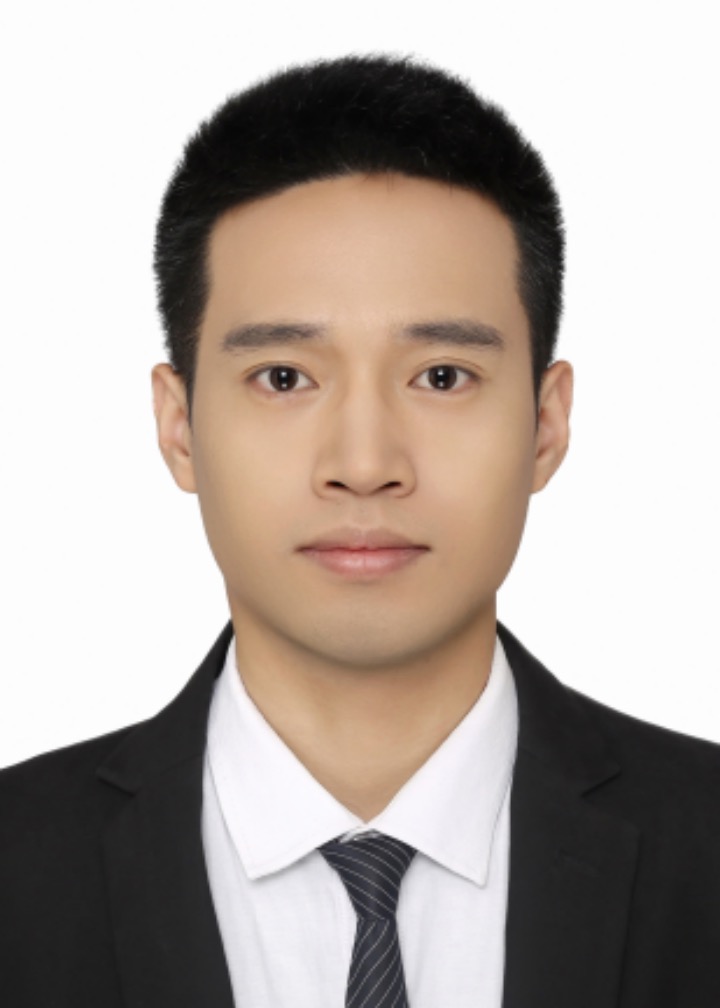}}]{Hui Luo} 
is currently pursuing a Ph.D. in signal and information processing at the University of Chinese Academy of Sciences in Beijing, China. During his Ph.D. program, he also conducts research at the Institute of Optics and Electronics, Chinese Academy of Sciences, located in Chengdu, China. His research interests include model compression and acceleration, as well as developing robust and reliable models.
\end{IEEEbiography}
\vskip -0.3in

\begin{IEEEbiography}[{\includegraphics[width=1in,height=1.25in,clip,keepaspectratio]{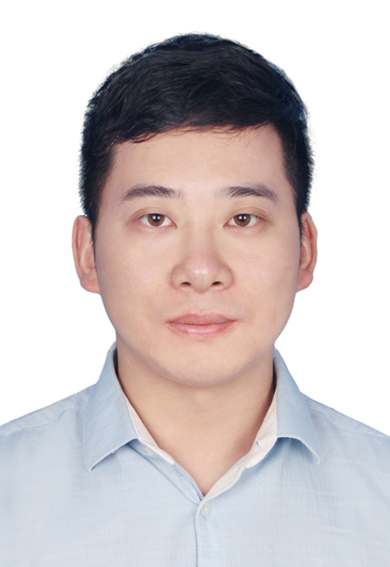}}]{Mingkui Tan}
is currently a Professor at the School of Software Engineering, South China University of Technology, Guangzhou, China. He received the Bachelor Degree in Environmental Science and Engineering in 2006 and the Master Degree in Control Science and Engineering in 2009, both from Hunan University in Changsha, China. He received the Ph.D. degree in Computer Science from Nanyang Technological University, Singapore, in 2014. From 2014-2016, he worked as a Senior Research Associate on computer vision at the School of Computer Science, University of Adelaide, Australia. His research interests include machine learning, sparse analysis, deep learning, and large-scale optimization.
\end{IEEEbiography}

\end{document}